%% file: main.tex
\documentclass[nohyperref]{article}

\usepackage{microtype}
\usepackage{graphicx}
\usepackage{subfigure}
\usepackage{booktabs} 

\usepackage{hyperref}

\usepackage[accepted]{icml2023}

\usepackage{enumerate}
\usepackage{comment}
\usepackage[utf8]{inputenc} 
\usepackage[T1]{fontenc}    
\usepackage{url}            
\usepackage{amsfonts}       
\usepackage{nicefrac}       
\usepackage{xcolor}         

\usepackage{amsmath, amsthm, amssymb}
\usepackage{bm}

\usepackage{tikz}
\usetikzlibrary{patterns}
\usetikzlibrary{automata, positioning}
\usetikzlibrary{shapes,decorations,arrows,calc,arrows.meta,fit,positioning}
\tikzset{
    -Latex,auto,node distance =1 cm and 1 cm,semithick,
    state/.style ={circle, draw, minimum width = 0.8 cm},
    state2/.style ={circle, draw, minimum width = 0.1 cm, inner sep=0pt},
    point/.style = {circle, draw, inner sep=0.04cm,fill,node contents={}},
    bidirected/.style={Latex-Latex,dashed},
    el/.style = {inner sep=2pt, align=left, sloped}
}
\usepackage{thmtools}
\usepackage{thm-restate}
\usepackage[normalem]{ulem}

\icmltitlerunning{Counterfactual Analysis in Dynamic Latent-State Models}

\input{definitions.tex}

\graphicspath{ {images/} }

\newcommand*{\Scale}[2][4]{\scalebox{#1}{$#2$}}%

\begin{document}

\twocolumn[
\icmltitle{Counterfactual Analysis in Dynamic Latent-State Models}

\icmlsetsymbol{equal}{*}

\begin{icmlauthorlist}
\icmlauthor{Martin Haugh}{AffImperial}
\icmlauthor{Raghav Singal}{AffDartmouth}
\end{icmlauthorlist}

\icmlaffiliation{AffImperial}{Imperial College}
\icmlaffiliation{AffDartmouth}{Dartmouth College}

\icmlcorrespondingauthor{MH}{m.haugh@imperial.ac.uk}
\icmlcorrespondingauthor{RS}{singal@dartmouth.edu}

\icmlkeywords{Causal inference,  state-space models, copulas, optimization, Monte-Carlo}

\vskip 0.3in
]

\printAffiliationsAndNotice{}

\begin{abstract}
We provide an optimization-based framework to perform counterfactual analysis in a dynamic model with hidden states.
Our framework is grounded in the ``abduction, action, and prediction'' approach to answer counterfactual queries and handles two key challenges where (1) the states are \emph{hidden} and (2) the model is \emph{dynamic}.
Recognizing the lack of knowledge on the underlying causal mechanism and the possibility of infinitely many such mechanisms,  we optimize over this space and compute  upper and lower bounds on the counterfactual quantity of interest.
Our work brings together ideas from causality,  state-space models,  simulation, and optimization,  and we apply it on a breast cancer case study. To the best of our knowledge, we are the first to compute lower and upper bounds on a counterfactual query in a dynamic latent-state model.
\end{abstract}

\section{Introduction} \label{sec:intro}

\emph{Counterfactual analysis}, falling on the third rung of Pearl's ladder of causation \cite{pearl_2018}, is a fundamental problem in causality.   It requires us to {\em imagine} a world where a certain policy was enacted with a corresponding outcome \emph{given} that a different policy  and outcome were actually observed.   It is performed via the 3-step framework of \emph{abduction} (conditioning on the observed data), \emph{action} (changing the policy),  and \emph{prediction} (computing the counterfactual quantity of interest (CQI)), and has wide-ranging applications \cite{pearl2009causal, pearl_2009}.

As a concrete application in healthcare and legal reasoning, consider someone who recently died from breast cancer. The exact progression of her disease is unknown. What is known, however, is that over a period of time prior to her diagnosis, her insurance company adopted a strategy of denying her regular scans (e.g.,  mammograms) even though these scans should have been covered by her policy. Had these scans gone ahead, the cancer may have been found earlier and the patient's life saved. Now a court wants to know the probability that her life would have been saved had the routine scans been permitted.

On top of the challenges posed by standard counterfactual analysis, there are two that are particular to such a setting.  First, it's possible the underlying state of the patient (e.g., stage of cancer) is \emph{hidden / latent} and we only observe a noisy signal depending on the accuracy of the scan (e.g., sensitivity and specificity of a mammogram).  Second, the underlying model is \emph{dynamic} as the patient's state evolves over time.  As such, our goal in this work is to perform counterfactual analysis in dynamic latent-state models.\footnote{The key feature distinguishing a static model from a dynamic model with $T$ periods say, is that the single-period structure is repeated $T$ times. As we shall see, our framework takes advantage of this repeated structure in several ways.}

Two streams of work are closely related to ours. The first relates to works on constructing bounds on CQIs \cite{BALKE-Pearl94,TianPearl2000,Kaufman2005,CaiEtAl2008,pearl_2009,Mueller_Li_Pearl_2021, Zhang-Tian-Barenboim-2021}.  These papers focus on \emph{static} models.
We note that despite some similarities of our work with \citet{Zhang-Tian-Barenboim-2021},  the two approaches are quite different. In particular, while both papers recognize the relevance of polynomial optimization for bounding CQIs, \citet{Zhang-Tian-Barenboim-2021} do not solve polynomial optimization problems but instead propose Monte-Carlo algorithms as a work-around. In contrast, we actually solve polynomial optimization problems  via sample average approximations (SAAs),  which we generate via Monte-Carlo. As such, Monte-Carlo serves as an ``input'' to our polynomial programs whereas \citet{Zhang-Tian-Barenboim-2021} use it as a ``substitute'' for polynomial programs. As mentioned above, another difference is our focus on dynamic models whereas \citet{Zhang-Tian-Barenboim-2021} focus on static models.
The second stream is more recent and concerns counterfactual analysis in dynamic models \cite{buesing2018, ICML209-Obers, NEURIPS2021_Lorberbom,NEURIPS2021_Tsirtsis}.  Except \citet{buesing2018}, none of these works allows for \emph{latent} states.  In addition,  these works perform counterfactual analysis by embedding assumptions that are strong enough to restrict the underlying set of causal mechanisms to a singleton.  In particular, \citet{buesing2018} explicitly fix a single causal mechanism whereas \citet{ICML209-Obers} and \citet{NEURIPS2021_Tsirtsis} invoke \emph{counterfactual stability} 
and implicitly fix the causal mechanism (via the Gumbel-max distribution). \citet{NEURIPS2021_Lorberbom} extend the Gumbel-max approach but their choice of causal mechanism is the one that minimizes variance when estimating the CQI. In summary, none of these approaches explicitly account for all possible causal mechanisms,  and therefore, they do not consider the construction of lower and upper bounds on the CQI, which is our focus.

Our key contribution is to provide a principled framework for counterfactual analysis in dynamic latent-state models.  We define our problem in \S\ref{sec:problem} and discuss counterfactual stability in \S\ref{sec:existing}.  In \S\ref{sec:opt},  we present our solution approach and we describe our numerics in \S\ref{sec:exp}.  We conclude in \S\ref{sec:conc}.

\section{Problem Definition} \label{sec:problem}
We first define the underlying dynamic latent-state model (\S\ref{sec:model}) and then describe the counterfactual analysis problem (\S\ref{sec:question}).
We will use a breast cancer application as a vehicle for explaining ideas throughout but it should be clear our framework is quite general.

 \subsection{A Dynamic Latent-State Model} \label{sec:model}

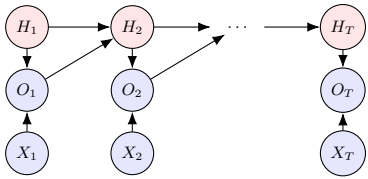
\begin{figure}[ht]
\centering
\begin{tikzpicture}[scale=.7,transform shape]
    \node[state,fill=red!10!white] (h1) at (0,0) {\footnotesize $H_1$};
    \node[state,fill=red!10!white] (h2) at (2,0) {\footnotesize $H_2$};
    \node[draw=none, minimum width = 1 cm] (hdots) at (4,0) {\footnotesize $\ldots$};
    \node[state,fill=red!10!white] (hT) at (6,0) {\footnotesize $H_T$};
    \node[state,fill=blue!10!white] (o1) at (0,-1.2) {\footnotesize $O_1$};
    \node[state,fill=blue!10!white] (o2) at (2,-1.2) {\footnotesize $O_2$};
    \node[state,fill=blue!10!white] (oT) at (6,-1.2) {\footnotesize $O_T$};
    \node[state,fill=blue!10!white] (x1) at (0,-2.4) {\footnotesize $X_1$};
    \node[state,fill=blue!10!white] (x2) at (2,-2.4) {\footnotesize $X_2$};
    \node[state,fill=blue!10!white] (xT) at (6,-2.4) {\footnotesize $X_T$};
    \path (h1) edge (o1);
    \path (h1) edge (h2);
    \path (o1) edge (h2);
    \path (h2) edge (o2);
    \path (h2) edge (hdots);
    \path (o2) edge (hdots);
    \path (hT) edge (oT);
    \path (hdots) edge (hT);
    \path (x1) edge (o1);
    \path (x2) edge (o2);
    \path (xT) edge (oT);
\end{tikzpicture}
\caption{A dynamic latent-state model.  States $H_{1:T}$ are hidden (red). Emissions $O_{1:T}$ are observed (blue).  $X_{1:T}$ represents the policy (observed).  (We use the notation $X_{1:T} := (X_t)_{t=1}^T$.) }
\label{fig:model}
\end{figure}

The model, visualized in Figure \ref{fig:model}, has $T$ discrete periods.  In each period $t$, the system is in a hidden state $H_t \in \bH$ (finite). As a stochastic function of $H_t$ and the policy $X_t \in \bX$ (finite),  we observe an emission $O_t \in \bO$ (finite). The \emph{emission probability} is denoted by $e_{hxi} := \bP(O_{t} = i \mid H_t = h, X_t=x)$ for all $(h,x,i)$. This is followed by the state $H_t$ transitioning to $H_{t+1}$ with \emph{transition probability} $q_{hih'} := \bP(H_{t+1} = h' \mid H_t = h, O_t=i)$.
The model $\fM$ comprises three primitives: $\fM \equiv (\fp,  \fE,  \fQ)$, where
$\fp := [p_h]_h$ denotes the \emph{initial state distribution} with $p_{h} := \Pb(H_1 = h)$ for all $h$,  $\fE := [e_{hxi}]_{h,x,i}$, and $\fQ := [q_{hih'}]_{h,i,h'}$.

In the breast cancer application,  the time periods map to the frequency of mammograms (e.g., 6 months) and the hidden state $H_t \in \{1,\ldots,7\}$ denotes the patient's condition. State 1 equates to the patient being healthy whereas states 2 and 3 correspond to \emph{undiagnosed} in-situ and invasive breast cancer, respectively. States 4 and 5 correspond to \emph{diagnosed} in-situ and invasive breast cancer respectively, with the understanding that the cancer treatment has begun (since it has been diagnosed). States 6 and 7 are absorbing and denote recovery from cancer (due to treatment) and  death from cancer, respectively.
The observation $O_t \in \{1,\ldots,7\}$ captures the mammogram result. A value of 1 means no screening took place,  whereas 2 denotes a negative screening result (possibly a false negative).  A value of 3 corresponds to a positive mammogram result, but followed by a negative biopsy (i.e., the patient is healthy and the mammogram produced a false positive).
Observations 4 and 5 map to correctly diagnosed in-situ and invasive cancer respectively, i.e., a positive mammogram followed by a positive biopsy.  Observations 6 and 7 are used to denote patient recovery and death from breast cancer, respectively.
The variable $X_t \in \{0,1\}$ models the insurance company's coverage policy for the mammograms, with 0 denoting the company covers it and 1 denoting the company (incorrectly) denies the coverage.  If the coverage is denied, then the mammogram is not performed and hence, the observation cannot be 2, 3, 4, or 5. (In this application, the $X_t$'s are deterministic but in general, they could be the result of a randomized policy.)

Our model is therefore a generalization of a hidden Markov model (HMM) since $H_{t+1}$ depends not only on $H_t$ but also on $O_t$.
The dependence on $O_t$ is needed to capture the fact that if cancer was detected during period $t$ and treatment began at that point of time, i.e., $O_t \in \{4,5\}$, then $H_{t+1}$ depends on the fact that the treatment began in period $t$. For example, if $H_t = 2$ (in-situ cancer) and $O_t = 4$ (in-situ diagnosed and hence, treatment began), then $H_{t+1}$ would be different compared to when $H_t=2$ and $O_t=2$ (false negative and hence, treatment did not begin).

\begin{rem} \label{rem:POMDP}
\citet{ayer2012or} employed a similar model for determining an optimal screening strategy for breast cancer but as their goal was to optimize over screening strategies, their model was a partially observable Markov decision process (POMDP). In contrast, our goal is not to find an optimal strategy but to evaluate CQIs. As such, our model is not a POMDP although it is easily related to a POMDP setting. For example, we can view the insurance company's observed coverage strategy and the counterfactual strategy where coverage is always provided,  as being feasible strategies from a POMDP. Finally, we also note that a  practical justification for our model comes from the simulation model used by the National Cancer Institute \citep{2013UWBCS}.
\end{rem}


\subsection{The Counterfactual Analysis Problem} \label{sec:question}

We now use our dynamic model to state the counterfactual analysis problem.

\paragraph{Observed data.}Suppose we observe emissions $o_{1:T}$ with the underlying policy being $x_{1:T}$.  The true hidden states $h_{1:T}$ are not observed.  In the context of breast cancer,  the observations for a particular patient might be as follows:
\begin{align}
\underbrace{o_1,  \ldots,  o_{\tau_s-1}}_{\in \{2,3\}},   \red{\underbrace{o_{\tau_s}, \ldots, o_{\tau_e}}_{=1}},  \underbrace{o_{\tau_e+1}, \ldots, o_{\tau_d-1}}_{\in \{4,5\}},  \underbrace{o_{\tau_d:T}}_{=7}.
\label{eq:path}
\end{align}
That is, the patient was screened ($x_{1:\tau_s-1}=0$) and appeared healthy ($o_{1:\tau_s-1} \in \{2,3\}$) up to and including time $\tau_s - 1$. Coverage was  denied during periods $\tau_s$ to $\tau_e$, i.e. $x_{\tau_s:\tau_e}=1$; (see red font in (\ref{eq:path}).  Hence, screening was not performed during those periods ($o_{\tau_s:\tau_e}=1$). As soon as the coverage for screening was re-approved (period $\tau_e + 1$ and hence, $x_{\tau_e+1} = 0$), the patient was found to have cancer (either in-situ or invasive) and the corresponding treatment began; thus, $o_{\tau_e + 1} \in \{4,5\}$. Unfortunately, the patient died at $\tau_d$.

\paragraph{CQI.} We focus on the well-known \emph{probability of necessity} (PN) \citep{pearl2009causal} as our CQI.  It is the probability the patient would have not died (counterfactual state $\tH_T \neq 7$) had the screening been covered in every period (intervention policy $\tx_{1:T} = 0$)  given the observed data $(o_{1:T}, x_{1:T})$.  (``Tilde'' notation denotes quantities in the counterfactual world.) The interpretation of $\tx_{1:T}$ is straightforward as it is fixed exogenously.  The counterfactual state $\tH_T$ is obtained via the 3 steps of abduction, action, and prediction \citep{pearl_2009}.  Step 1 (abduction) involves conditioning on the observed data $(o_{1:T}, x_{1:T})$ to form a posterior belief over the hidden states.  Step 2 (action) changes the policy from $x_{1:T}$ to $\tx_{1:T}$ and brings us to the counterfactual world $\tM$.
Step 3 (prediction) computes PN in the counterfactual model:
\begin{align}
\text{PN} = \Pb(\tH_T \neq 7),
\label{eq:PN}
\end{align}
with the understanding that the event $\{\tH_T \neq 7\}$ is conditional on $(o_{1:T}, x_{1:T})$.
Though we focus on PN,  it is easy to extend our framework to a broad class of CQIs as the abduction and action steps do not depend on the CQI. 

Given our focus on counterfactual analysis, we will assume the primitives $(\fp,  \fE,  \fQ)$ are known.  We discuss their calibration to real-world data in  \S\ref{sec:exp} and emphasize that even with known $(\fp,  \fE,  \fQ)$, counterfactual analysis is challenging.  This is because we are interested in counterfactuals at an \emph{individual} level (i.e., conditioning on the patient-level data $(o_{1:T}, x_{1:T})$ via abduction), as opposed to the \emph{population} level. A population-level counterfactual analysis would ignore the first step of abduction but simply change the policy to $\tx_{1:T}$ to predict the CQI (by simulating the resulting model and obtaining a Monte-Carlo estimate of PN or doing it in closed-form if analytically tractable). However, this is very different from the task at hand, which falls on the highest rung of Pearl's ladder of causation \cite{pearl_2018}.  For instance, consider a patient who dies immediately after the coverage was denied versus a patient who dies a couple of years after the coverage was denied. Clearly, the first patient had a more ``aggressive'' cancer and hence we expect that her PN would be lower.  By conditioning on individual-level data $(o_{1:T}, x_{1:T})$,  we are able to account for such differences.  However, it makes the problem considerably more challenging.

In our dynamic latent-state model, each of the three steps of abduction, action, and prediction presents its own set of challenges\footnote{Instead of using the ``twin networks'' approach \citep{pearl_2009}, we perform the counterfactual analysis directly by leveraging the structure in our model.}, which we discuss when presenting our methodology in \S\ref{sec:opt}.   Before doing so, we discuss the notion of counterfactual stability (CS), which has become a popular approach in some settings \citep{ICML209-Obers}.

\section{Limitations of Counterfactual Stability}  \label{sec:existing}

Instead of discussing CS in our dynamic latent-state model, we do so using the following simple model: $X \to Y$. 
Suppose we observe an outcome $Y=y$ under policy $X=x$.
With  $Y_x := Y \mid (X = x)$,
CS requires that the counterfactual outcome under an interventional policy $\tx$ (denoted by $\tY := Y_{\tx} \mid Y_x = y$) cannot be $y'$ (for $y' \not = y$) if $\bP(Y_{\tx}=y) / \bP(Y_x=y) \geq \bP(Y_{\tx}=y') / \bP(Y_x=y')$.  In words, CS states that if $y$ was observed and this outcome becomes relatively more likely than $y'$ under the intervention, then the counterfactual outcome $\tY$ can not be $y'$.


Though somewhat appealing,  the appropriateness of CS depends on the application and should  be justified by domain specific knowledge.  Moreover, we show in Example \ref{ex:CS} that CS can permit counterfactuals that it was seemingly designed to exclude.

\begin{ex} \label{ex:CS}
Consider the $X \to Y$ model and suppose $X \in \{0,1\}$ denotes a medical treatment and $Y \in \{\text{bad}, \text{better}, \text{best}\}$ the patient outcome.  For illustration, suppose the outcome $Y_x$ obeys the following distribution: $Y_0 \sim \{\text{bad}, \text{better}, \text{best}\}$ w.p.\ $\{0.2, 0.3, 0.5\}$ and $Y_1 \sim \{\text{bad}, \text{better}, \text{best}\}$ w.p.\ $\{0.2, 0.2, 0.6\}$.  That is, under treatment ($x=1$), the ``best'' outcome becomes more likely but the likelihood of the ``bad'' outcome does not change.
Consider a patient whose outcome $Y$ was ``better'' under no treatment ($x=0$).
Suppose also that domain specific knowledge tell us that even at the individual level, the counterfactual outcome $\tY$ should not be worse under treatment ($\tx=1$) than under no treatment  ($x=0$).   However,  since
\begin{align*}
\frac{\bP(Y_1=\text{better})}{\bP(Y_0=\text{better})} = \frac{0.2}{0.3} < \frac{0.2}{0.2} = \frac{\bP(Y_1=\text{bad})}{\bP(Y_0=\text{bad})},
\end{align*}
``bad'' is a feasible counterfactual outcome under CS.
\end{ex}

Even if CS is appropriate, its current operationalization has a key limitation.  In particular, instead of considering all possible structural causal models (SCMs) that obey CS, both \citet{ICML209-Obers} and \citet{NEURIPS2021_Tsirtsis} pick one SCM via the Gumbel-max distribution. Ideally, one should characterize the space of all SCMs obeying CS, and map that space into appropriate bounds on the CQI.

We present our optimization-based framework to perform counterfactual analysis next.
Our framework does not rely on CS.  However, \emph{if} CS is deemed appropriate for one or more components of the SCM (see \S\ref{sec:opt}), our approach allows us to encode CS via linear constraints in the optimization and characterize the \emph{entire} space of solutions that obey CS.  We do this in \S\ref{sec:exp} to negatively answer the open question of \citet{ICML209-Obers} regarding whether Gumbel-max obeys CS uniquely.
Further,  if enforcing the so-called \emph{pathwise monotonicity} (PM) is desirable, i.e.,  ensuring the counterfactual outcome does not worsen under a better intervention (as we assumed in Example \ref{ex:CS}), then we can embed it in our optimization via linear constraints as well.

\section{Counterfactual Analysis via Optimization}  \label{sec:opt}
We now present our solution methodology for the counterfactual analysis problem introduced in \S\ref{sec:problem}.
We first discuss the underlying SCM (\S\ref{sec:SCM}), which is a precursor to defining the counterfactual model $\tM$ (\S\ref{sec:CFModel}), which feeds into our optimization framework for counterfactual analysis (\S\ref{sec:optProblem}).

\subsection{The Structural Causal Model (SCM)}  \label{sec:SCM}

\begin{figure}[ht]
\centering
\begin{tikzpicture}[scale=.7,transform shape]
    \node[state,fill=red!10!white] (h1) at (0,0) {\footnotesize $H_1$};
    \node[state,fill=red!10!white] (h2) at (2,0) {\footnotesize $H_2$};
    \node[state2,fill=black!10!white] (u2) at (1,1) {\footnotesize $\fU_2$};
    \node[draw=none, minimum width = 1 cm] (hdots) at (4,0) {\footnotesize $\ldots$};
    \node[state,fill=red!10!white] (hT) at (6,0) {\footnotesize $H_T$};
    \node[state2,fill=black!10!white] (uT) at (5,1) {\footnotesize $\fU_T$};
    \node[state,fill=blue!10!white] (o1) at (0,-1.2) {\footnotesize $O_1$};
    \node[state2,fill=black!10!white] (v1) at (-1,-1.8) {\footnotesize $\fV_1$};
    \node[state,fill=blue!10!white] (o2) at (2,-1.2) {\footnotesize $O_2$};
    \node[state2,fill=black!10!white] (v2) at (1,-1.8) {\footnotesize $\fV_2$};
    \node[state,fill=blue!10!white] (oT) at (6,-1.2) {\footnotesize $O_T$};
    \node[state2,fill=black!10!white] (vT) at (5,-1.8) {\footnotesize $\fV_T$};
    \node[state,fill=blue!10!white] (x1) at (0,-2.4) {\footnotesize $X_1$};
    \node[state,fill=blue!10!white] (x2) at (2,-2.4) {\footnotesize $X_2$};
    \node[state,fill=blue!10!white] (xT) at (6,-2.4) {\footnotesize $X_T$};
    \path (h1) edge (o1);
    \path (h1) edge (h2);
    \path (o1) edge (h2);
    \path (h2) edge (o2);
    \path (h2) edge (hdots);
    \path (o2) edge (hdots);
    \path (hT) edge (oT);
    \path (hdots) edge (hT);
    \path (x1) edge (o1);
    \path (x2) edge (o2);
    \path (xT) edge (oT);
    \path (u2) edge (h2);
    \path (uT) edge (hT);
    \path (v1) edge (o1);
    \path (v2) edge (o2);
    \path (vT) edge (oT);
\end{tikzpicture}
\caption{The SCM underlying the dynamic latent-state model.  The only difference between the SCM here and Figure \ref{fig:model} is the addition of (grey) exogenous noise nodes $[\fU_t, \fV_t]_t$. }
\label{fig:SCM}
\end{figure}
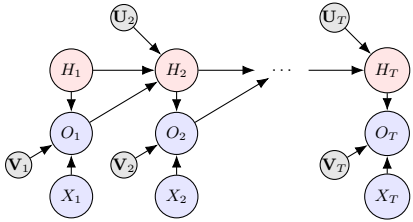

To understand the SCM (Figure \ref{fig:SCM}),  consider $O_t$ for any $t$, which is a stochastic function of its parents $(H_t, X_t)$. The stochasticity is driven by the exogenous noise vector $\fV_t := [V_{thx}]_{h,x}$, which comprises of $\lvert\bH\rvert \lvert \bX\rvert$ noise variables.
We model the exogenous node as a vector (as opposed to a scalar) to capture the fact that each $O_{thx} := O_t \mid (H_t = h, X_t = x)$ defines a \emph{distinct} random variable for all $(h,x)$.  Moreover, these random variables might be independent, or they might display positive or negative dependence.
One way to handle this is to associate each $O_{thx}$ with a distinct noise variable $V_{thx}$.  The {\em dependence structure} among these noise variables $[V_{thx}]_{h,x}$ is then what determines the dependence structure among $[O_{thx}]_{h,x}$.  The structural equation obeys
\begin{subequations}
\label{eq:SCM}
\begin{align}
\Scale[1]{O_t = f(H_t, X_t, \fV_t) = \sum_{h,x} f_{hx}(V_{thx}) \bI_{\{H_t=h, X_t=x\}}}
\label{eq:SCMone}
\end{align}
where $f_{hx}(\cdot)$ is defined using the emission distribution $[e_{hxi}]_i$ and $V_{thx} \sim \text{Unif}[0,1]$ wlog.
Similarly,  for $t>1$, recognizing that each $H_{thi} := H_{t} \mid (H_{t-1} = h, O_{t-1} = i)$ is a distinct random variable for all $(h,i)$, we associate each $H_{thi}$ with its own noise variable $U_{thi}$:
\begin{align}
\Scale[1]{H_t = g(H_{t-1},  O_{t-1}, \fU_t) = \sum_{h,i} g_{hi}(U_{thi}) \bI_{\{H_{t-1}=h, O_{t-1}=i\}}}
\label{eq:SCMtwo}
\end{align}
\end{subequations}
where $g_{hi}(\cdot)$ is defined using the transition distribution $[q_{hih'}]_{h'}$ and $U_{thi} \sim \text{Unif}[0,1]$ wlog.

The representation in \eqref{eq:SCMone} allows us to model $[O_{thx}]_{h,x}$ and capture any dependence structure among these random variables by specifying the joint multivariate distribution of $\fV_t$.  Since the univariate marginals of $\fV_t$ are known ($\text{Unif}[0,1]$), specifying the multivariate distribution amounts to specifying the dependence structure or {\em copula}.  (Of course, the same comment applies to $\eqref{eq:SCMtwo}$ and $\fU_t$ as well.) For example, if the $V_{thx}$'s are mutually independent (the independence copula) and we have $(H_t=h', X_t=x')$,  then inferring the conditional distribution of $V_{th'x'}$ will tell us nothing about the $V_{thx}$'s for $(h,x) \neq (h',x')$.  Alternatively, if $V_{thx} = V_{th'x'}$ for all pairs $(h,x)$ and $(h',x')$, then this models perfect positive dependency (the comonotonic copula) and inferring the conditional distribution of $V_{th'x'}$ amounts to simultaneously inferring the conditional distribution of all the $V_{thx}$'s.
We emphasize that we must work with the exogenous vectors $(\fU_t,\fV_t)$ when doing a {\em counterfactual} analysis since different joint distributions of $(\fU_t,\fV_t)$ will lead to (possibly very) different values of PN. If we are not doing a counterfactual analysis and only care about the joint distribution of a (subset of) $(O_{1:T}, H_{1:T})$ then our analysis will only depend on the joint distribution of the $(\fU_t,\fV_t)$'s via their known univariate marginals.  We note the $\fU_t$'s and $\fV_t$'s must be mutually independent in order for the SCM to be consistent with the dependence / independence relationships implied by the model of Figure \ref{fig:model}.

In our dynamic model, the emissions and the state transitions are time-independent. Thus, it is natural to also assume the copulas underlying $\fV_t$ and $\fU_t$ are time-independent. We refer to this property as \emph{time invariance}.
As such,  we define the notation $O_{hx} := O_t \mid (H_t = h, X_t = x)$ and $H_{hi} := H_{t+1} \mid (H_t = h, O_t = i)$.\footnote{$O_t \mid (H_t = h, X_t = x)$ is time-independent and hence, we use the notation $O_{hx}$ instead of $O_{thx}$. Same logic holds for $H_{hi}$.} Then,  $e_{hxi} = \bP(O_{hx} = i)$ and $q_{hih'} = \bP(H_{hi} = h')$.

While the copula view is useful from a conceptual point of view (since specifying copulas for $\fU_t$ and $\fV_t$ amounts to specifying an SCM),  it is more convenient to work with an alternative construction of the SCM.  This is because in discrete-state space models, there will be infinitely many joint distributions of $\fV$ (and $\fU$) that all lead to the same joint distribution of $[O_{hx}]_{h,x}$ (and $[H_{hi}]_{h,i}$). In other words, the joint distribution of $[V_{hx}]_{h,x}$ does not uniquely identify the joint distribution of $[O_{hx}]_{h,x}$. This is a consequence of Sklar's Theorem from the theory of copulas and is discussed\footnote{In \S\ref{sec:CopulaAppendix}, we also discuss specific copulas (e.g., independence and comonotonic copulas) that can be used to provide benchmark values of PN.}  further in \S\ref{sec:CopulaAppendix}. We will therefore take a more direct approach by modeling the unknown joint distribution of $[O_{hx}]_{h,x}$ (and $[H_{hi}]_{h,i}$).
As such, we define
\begin{subequations}
\label{eq:PairwiseMarginals}
\begin{align}
\theta_{\tildeh \tx, hx}(\ti, i) &:= \Pb (O_{\tildeh \tx} = \ti,  O_{hx} = i ) \\
\pi_{\tildeh \ti, hi}(\tildeh', h') &:= \Pb (H_{\tildeh \ti} = \tildeh',  H_{hi} = h' )
\end{align}
\end{subequations}
and observe that
\begin{subequations}
\label{eq:marginalconstraintstwo}
\begin{align}
\theta_{hx,hx}(i,i) &= e_{hxi} \ \forall (h,x,i) \\
\pi_{hi,hi}(h',h') &= q_{hih'} \ \forall (h,i,h').
\end{align}
\end{subequations}
This holds because $\pi_{hi,hi}(h',h')  = \Pb (H_{hi} = h',  H_{hi} = h' ) =\Pb (H_{hi} = h') = q_{hih'}$.
We also have \emph{symmetry}, i.e.,
\begin{subequations}
\label{eq:symmetry}
\begin{align}
\theta_{\tildeh \tx,  hx}(\ti, i) &= \theta_{hx, \tildeh \tx}(i,\ti) \ \forall (\tildeh, \tx, \ti) \ \forall (h,x,i) \\
\pi_{\tildeh \ti,  h i}(\tildeh',h') &= \pi_{h i, \tildeh \ti}(h', \tildeh') \ \forall (\tildeh,\ti,\tildeh') \ \forall (h,i,h').
\end{align}
\end{subequations}
This is because $\pi_{\tildeh \ti,  h i}(\tildeh',h') = \Pb (H_{\tildeh \ti} = \tildeh',  H_{hi} = h' ) = \Pb (H_{hi} = h',  H_{\tildeh \ti} = \tildeh') = \pi_{h i, \tildeh \ti}(h', \tildeh')$.
We only defined the ``pairwise marginals'' in \eqref{eq:PairwiseMarginals} but we will define the full joint PMFs in \eqref{eq:JointVariables}. We are now ready to discuss the counterfactual model.

\subsection{The Counterfactual Model $\tM$}  \label{sec:CFModel}
Recall from \S\ref{sec:problem} that $\tM$ is obtained after the two steps of abduction (conditioning on the observed data $(o_{1:T}, x_{1:T})$) and action (changing the policy from $x_{1:T}$ to $\tx_{1:T}$).
Understanding the dynamics underlying $\tM$ are non-trivial, primarily due to the abduction step where the goal is to obtain the posterior distribution of the hidden path $H_{1:T}$. It is not possible to provide a closed-form expression for this distribution  but we can use filtering / smoothing methods to describe the posterior dynamics of $H_{1:T}$. (See \S\ref{sec:sampling} for details.)

We can therefore use these dynamics to generate $B$ Monte-Carlo samples $[h_{1:T}(b)]_{b=1}^B$ from the posterior, i.e., from the distribution of $H_{1:T} \mid (o_{1:T}, x_{1:T})$.  Then, by conditioning on each sample $b$, it is possible to characterize  $\tM$.
In particular, denote by $\tM(b) \equiv (\tfp(b),  [\tfE^{(t)}(b)]_t,  [\tfQ^{(t)}(b)]_t)$ the counterfactual model corresponding to posterior sample $h_{1:T}(b)$.
Similar to the primitives $(\fp,  \fE,  \fQ)$ in \S\ref{sec:problem}, the counterfactual primitives $(\tfp(b),  [\tfE^{(t)}(b)]_t,  [\tfQ^{(t)}(b)]_t)$ correspond to initial state, emission, and transition distributions.
As $H_1$ in Figure \ref{fig:model} has no parents, $\tfp(b)$ is such that the counterfactual hidden state in period 1 equals the posterior sample $h_1(b)$, i.e., $\tildeh_1(b) = h_1(b)$.  In contrast with $\fE$ and $\fQ$, both $\tfE^{(t)}(b)$ and $\tfQ^{(t)}(b)$ are time-dependent (note the super-script ``$(t)$'').
This is because the period $t$ counterfactual emission $\tfE^{(t)}(b) := [\te^{(t)}_{\tildeh \tx \ti}(b)]_{\tildeh,\tx,\ti}$ and transition $\tfQ^{(t)}(b) := [\tq^{(t)}_{\tildeh \ti \tildeh'}(b)]_{\tildeh,\ti, \tildeh'}$ probabilities are as follows:
\begin{subequations}
\label{eq:CFprimitives}
\begin{align}
\te^{(t)}_{\tildeh \tx \ti}(b) &= \Pb( O_{\tildeh \tx} = \ti \mid O_{h_t(b) x_t} = o_t ) \\
\tq^{(t)}_{\tildeh \ti \tildeh'}(b) &= \Pb( H_{\tildeh \ti} = \tildeh' \mid H_{h_t(b) o_t} = h_{t+1}(b) ).
\end{align}
\end{subequations}
(The $O_{hx}$ and $H_{hi}$ notation is defined above \eqref{eq:PairwiseMarginals}.)
The dependence on $t$ is through the observed data $(o_t, x_t)$ and the posterior samples $(h_t(b), h_{t+1}(b))$. As such, for each posterior path $b$, $\tM(b)$ is a time-dependent dynamic latent-state model.
If we knew $\tfE^{(t)}(b)$ and $\tfQ^{(t)}(b)$, then we could simulate $\tM(b)$ to obtain a Monte-Carlo estimate of our CQI by averaging the CQI over the $B$ posterior sample paths.
However,  $\tfE^{(t)}(b)$ and $\tfQ^{(t)}(b)$ are unknown as they depend on the joint distributions of $\fU_t$ and $\fV_t$.

Towards this end, we can combine \eqref{eq:CFprimitives} with \eqref{eq:PairwiseMarginals} to obtain
\begin{align*}
\te^{(t)}_{\tildeh \tx \ti}(b) &= \frac{\theta_{\tildeh \tx, h_t(b) x_t}(\ti,  o_t)}{e_{h_t(b) x_t o_t}} \\
\tq^{(t)}_{\tildeh \ti \tildeh'}(b) &= \frac{\pi_{\tildeh \ti, h_t(b) o_t}(\tildeh', h_{t+1}(b))}{q_{h_t(b) o_t h_{t+1}(b)}},
\end{align*}
which express the unknown and time-dependent emission and transition distributions in terms of the unknown $\ftheta$ and $\fpi$ that are  time-independent.

\subsection{Polynomial Optimization}  \label{sec:optProblem}

We now propose an optimization model where we treat the unknowns $(\ftheta, \fpi)$ as decisions and maximize (minimize) the CQI to obtain an upper bound (lower bound).
We present our optimization model in terms of the objective and constraints, followed by a discussion on how we can enforce CS and PM (if indeed they were deemed appropriate).

\paragraph{Objective.} As in \eqref{eq:PN}, we wish to understand the PN, which equals $\Pb(\tH_T \neq 7)$,  where $\tH_T$ is the hidden state at time $T$ under $\tM$.  The randomness in $\tH_T$ depends on the randomness in (i) the true hidden path $H_{1:T}$ (captured by $[h_{1:T}(b)]_b$)  and (ii) the counterfactual model $\tM \mid H_{1:T}$ after conditioning on $H_{1:T}$ (captured by $\tM(b)$).
Lemma \ref{lem:PNdecomp} decomposes PN using these two uncertainties. (All proofs are in \S\ref{sec:proofs}.)

\begin{restatable}{lem}{PNdecomp}  \label{lem:PNdecomp}
We have $$\text{PN} = 1 - \lim_{B \to \infty}\frac{1}{B} \sum_{b=1}^B \Pb_{\tM(b)} (\tH_T = 7).$$
\end{restatable}

We next express $\Pb_{\tM(b)} (\tH_T)$ in terms of $(\ftheta, \fpi)$ from \eqref{eq:PairwiseMarginals}.

\begin{restatable}{lem}{PNrecursive}  \label{lem:PNrecursive}
For $t \in \{T,  T-1, \ldots, 2\}$, $\Pb_{\tM(b)} (\tildeh_t) := \Pb_{\tM(b)} (\tH_t = \tildeh_t)$ obeys the following recursion (over $t$):
\begin{align*}
\Pb_{\tM(b)} (\tildeh_t) = \sum_{\tildeh_{t-1}, \tildeo_{t-1}} & \frac{\pi_{\tildeh_{t-1} \tildeo_{t-1}, h_{t-1}(b) o_{t-1}}(\tildeh_t, h_{t}(b))}{q_{h_{t-1}(b) o_{t-1} h_{t}(b)}} \times \\ & \frac{\theta_{\tildeh_{t-1} \tx_{t-1}, h_{t-1}(b) x_{t-1}}(\tildeo_{t-1},  o_{t-1})}{e_{h_{t-1}(b) x_{t-1} o_{t-1}}} \times \\  & \Pb_{\tM(b)}(\tildeh_{t-1}).
\end{align*}
The recursion breaks at $t=1$:
\begin{align*}
\Pb_{\tM(b)} ( \tildeh_1 ) = \begin{cases}
1 &\text{ if } \tildeh_1 = h_1(b) \\
0 &\text{ otherwise.}
\end{cases}
\end{align*}
\end{restatable}
Putting together Lemmas \ref{lem:PNdecomp} and \ref{lem:PNrecursive} allows us to express PN in terms of the various primitives, all of which except $(\ftheta, \fpi)$ are known (or can be sampled).  Thus,  we use the notation $\text{PN}(\ftheta, \fpi \mid [h_{1:T}(b)]_b)$.
As soon as we fix $(\ftheta, \fpi)$, we can estimate PN. However, it is unclear apriori what we should fix $(\ftheta, \fpi)$ at.  We might have some information on the structure of $(\ftheta, \fpi)$ that can help us shrink their feasibility space but in general, there can be many $(\ftheta, \fpi)$s that are ``valid''.  To overcome this lack of knowledge,  we take an agnostic view and compute bounds over PN. The upper (lower) bound  is computed by maximizing (minimizing) $\text{PN}(\ftheta, \fpi \mid [h_{1:T}(b)]_b)$ over the set of $(\ftheta, \fpi)$ that are ``valid''.  Denoting by $\cF$ the set of ``valid'' $(\ftheta, \fpi)$ (discussed below), we define
\begin{restatable}{subequations}{MaxMin}
\label{eq:MaxMin}
\begin{align}
\PNUB(B) &:= \max_{(\ftheta, \fpi) \in \cF} \text{PN}(\ftheta, \fpi \mid [h_{1:T}(b)]_b) \\
\PNLB(B) &:= \min_{(\ftheta, \fpi) \in \cF} \text{PN}(\ftheta, \fpi \mid [h_{1:T}(b)]_b).
\end{align}
\end{restatable}
Both optimizations in \eqref{eq:MaxMin} are sample average approximations (SAA) due to the use of the Monte-Carlo samples $[h_{1:T}(b)]_b$. Thus, $\PNUB(B)$ and $\PNLB(B)$ are estimates of the ``true'' $\PNUB$ and $\PNLB$.  However,  given $[h_{1:T}(b)]_b$ are iid samples, the following consistency result is immediate (cf.\ Proposition 5.2 in \citet{shapiro2021lectures}).
\begin{restatable}{prop}{SAA} \label{prop:SAA}
$(\PNLB(B), \PNUB(B))$ converges to $(\PNLB, \PNUB)$ w.p.\ 1 as $B \to \infty$.
\end{restatable}
In addition, we can characterize the asymptotics of $(\PNLB(B), \PNUB(B))$ via results in the SAA theory and we refer the reader to \S5.1.2 of \citet{shapiro2021lectures}.

\paragraph{Constraints (feasibility set $\cF$).}
We now discuss the feasibility set $\cF$.  Recall from \eqref{eq:PairwiseMarginals} that we used $\ftheta$ and $\fpi$ to denote the pairwise marginal distributions  over $[O_{hx}]_{h,x}$ and $[H_{hi}]_{h,i}$ respectively. We will now also use them to represent the full joint distributions of $[O_{hx}]_{h,x}$ and $[H_{hi}]_{h,i}$ respectively. To simplify notation, let $k \equiv (h,x)$ and $m \equiv (h,i)$. Hence,
\begin{align*}
O_{k} &\equiv O_{hx}, \ e_{k i} \equiv e_{hxi}  \\
H_m &\equiv H_{hi}, \ q_{mh'} \equiv q_{hih'}.
\end{align*}
We have $k \in [K]$ and $m \in [M]$, where $K := \lvert\bH\rvert \lvert \bX\rvert$ and $M := \lvert\bH\rvert \lvert \bO\rvert$.  The $K$ and $M$ dimensional joint PMFs for all $i_1, \ldots, i_K \in \bO$ and $h_1, \ldots, h_M \in \bH$ are defined as
\begin{restatable}{subequations}{JointVariables}
\label{eq:JointVariables}
\begin{align}
\theta_{1,\ldots,K}(i_1, \ldots, i_K) &:= \Pb (O_1 = i_1,  \ldots, O_K = i_K )  \\
\pi_{1,\ldots,M}(h_1, \ldots, h_M) &:= \Pb (H_1 = h_1,  \ldots, H_M = h_M ).
\end{align}
\end{restatable}
Note that we only have \emph{one} joint $\theta_{1,\ldots,K}$ among $K$ random variables in contrast to \emph{multiple} pairwise marginals $[\theta_{k \ell}]_{(k,\ell)}$.  Each of these joint PMFs are decision variables in the optimization (in addition to the pairwise decision variables) and must obey the following set of constraints. First,  the 1-dimensional marginals of $\ftheta$ and $\fpi$ must equal the given 1-dimensional marginals $[e_{ki}]_{(k,i)}$ and $[q_{mh}]_{(m,h)}$:
\begin{subequations}
\label{eq:marginalconsthetapi}
\begin{align}
\sum_{\{i_1, \ldots, i_K\} \setminus \{i_{k}\}} \theta_{1,\ldots,K}(i_1, \ldots, i_K) &= e_{1i_{k}} \ \forall i_{k} \in \bO, k \in [K] \label{eq:marginalconstheta} \\
\sum_{\{h_1,\ldots,h_M\} \setminus \{h_{m}\}}  \pi_{1,\ldots,M}(h_1, \ldots, h_M) &= q_{1h_{m}} \ \forall h_{m} \in \bH, m \in [M]. \label{eq:marginalconspi}
\end{align}
\end{subequations}
Recall that $(\fQ, \fE)$, i.e., the right-hand-sides of (\ref{eq:marginalconsthetapi}), are known.  Moreover, since  $\fQ$ and $\fE$ themselves define 1-dimensional probability distributions and therefore sum to 1, (\ref{eq:marginalconsthetapi}) ensures the same will be true of both the joint PMFs, i.e., they will also sum to 1.
Second, we must link the pairwise marginals to the joints:
\begin{subequations}
\label{eq:pairwisemarginallink}
\begin{align}
\theta_{k \ell}(i_k, i_{\ell}) &= \sum_{\{i_1, \ldots, i_K\} \setminus \{i_k, i_{\ell}\}} \theta_{1,\ldots,K}(i_1, \ldots, i_K)  \label{eq:pairwisemarginallinktheta} \\
\pi_{m n}(h_m, h_n) &= \sum_{\{h_1, \ldots, h_M\} \setminus \{h_m,  h_n\}} \pi_{1,\ldots,M}(h_1, \ldots, h_M).   \label{eq:pairwisemarginallinkpi}
\end{align}
\end{subequations}
\eqref{eq:pairwisemarginallinktheta} holds for all $i_k, i_{\ell} \in \bO$ and $k, \ell \in [K] \text{ s.t. } k < \ell$ whereas \eqref{eq:pairwisemarginallinkpi} holds for all $h_m, h_n \in \bH$ and $m,  n \in [M] \text{ s.t. } m < n$.  The ``$k < \ell$'' and ``$m < n$'' conditions avoid unnecessary duplication (recall \eqref{eq:marginalconstraintstwo} and \eqref{eq:symmetry}).\footnote{In fact, given \eqref{eq:marginalconstraintstwo} and \eqref{eq:symmetry}, we do not need to define all pairwise marginals as decision variables but only for ``$k < \ell$'' and ``$m < n$''.  This is because if an optimization has two decision variables $x$ and $y$ and the constraint $x=y$, we can eliminate $y$ and the constraint $x=y$ by replacing $y$ with $x$ everywhere in the optimization. \label{ft:elimination}}
Finally, we need to ensure non-negativity:
\begin{align}
\label{eq:nonnegativityjoint}
\ftheta, \fpi \ge 0
\end{align}
where we now use $(\ftheta, \fpi)$ to denote all of the corresponding, i.e., joint and pairwise, decision variables.

Let $\cF$ be the feasible region over $(\ftheta, \fpi)$ defined by the constraints \eqref{eq:marginalconsthetapi}, \eqref{eq:pairwisemarginallink},  and \eqref{eq:nonnegativityjoint}.
Observe that $\text{PN}(\ftheta, \fpi)$ is a polynomial in $(\ftheta, \fpi)$ (cf.\ Lemmas \ref{lem:PNdecomp} and \ref{lem:PNrecursive}) and the constraints in $\cF$ are linear.  Thus,  each of the problems in \eqref{eq:MaxMin} fall within the class of polynomial optimization \citep{anjos2011handbook}.
Denoting by $\text{PN}^*$ the PN under the true (unknown) $(\ftheta, \fpi)$, we obtain the following inequalities.

\begin{restatable}{prop}{PNBounds} \label{prop:PNBounds}
$\PNLB \le \text{PN}^* \le \PNUB.$
\end{restatable}

\paragraph{Enforcing CS and PM via linear constraints.} Suppose that at some time, the patient was in state $h$, the emission was $i$,  followed by a transition to state $h'$.  This maps to the realization $H_{hi} = h'$.  For $\tildeh' \neq h'$, CS requires that if $\bP(H_{\tildeh \ti} = h') / \bP(H_{hi} = h') \ge \bP(H_{\tildeh \ti} = \tildeh' ) / \bP(H_{hi} = \tildeh' )$, then $\bP(H_{\tildeh \ti} = \tildeh'  \mid H_{hi} = h') = 0$. Observe that the ``if'' condition is equivalent to $q_{\tildeh \ti h'} / q_{hih'} \ge q_{\tildeh \ti \tildeh' } / q_{hi\tildeh' }$ and the LHS of ``then'' equals $\pi_{\tildeh \ti, h i}(\tildeh' ,h')/q_{hih'}$. Hence, for the state transitions, CS is equivalent to adding the following linear constraints for all $(h,i,h', \tildeh, \ti, \tildeh' )$:
\begin{subequations}
\label{eq:CS}
\begin{align}
\pi_{\tildeh \ti, hi}(\tildeh' ,h') = 0 \text{ if } \frac{q_{\tildeh \ti h'}}{q_{hih'}} \ge \frac{q_{\tildeh \ti \tildeh' }}{q_{hi \tildeh' }}.
\label{eq:CSOne}
\end{align}
Similarly, for emissions,  CS can be modeled by adding the following linear constraints for all $(h,x,i,\tildeh,\tx,\ti)$:
\begin{align}
\theta_{\tildeh \tx, hx}(\ti,i) = 0 \text{ if } \frac{e_{\tildeh \tx i}}{e_{hxi}} \ge \frac{e_{\tildeh \tx \ti}}{e_{hx \ti}}.
\label{eq:CSTwo}
\end{align}
\end{subequations}
Hence,  we can characterize the space of all SCMs that obey CS, which is in contrast to picking just one such SCM \citep{ICML209-Obers}. Enforcing CS naturally leads to tighter bounds, but the bounds may not be ``legitimate'' if the true $(\ftheta, \fpi)$ does not satisfy CS.  Denoting by $\PNUBCS$ and $\PNLBCS$ the bounds obtained by adding CS constraints \eqref{eq:CS} to the optimizations in \eqref{eq:MaxMin}, we have the following result.
\begin{restatable}{prop}{CSBounds} \label{prop:CSBounds}
$\PNLB \le \PNLBCS \le \PNUBCS \le \PNUB.$
\end{restatable}

PM can also be enforced via linear constraints. To see this,  suppose the patient has in-situ cancer in period $t$ which is not detected but the patient's state remains at in-situ in period $t+1$. Then, in the counterfactual world, if the cancer is detected in period $t$, then PM would require that the cancer can not be worse than in-situ in period $t+1$, i.e.,
\begin{align*}
\bP(H_{\tildeh \ti} = \tildeh' \mid H_{hi} = h') = 0
\end{align*}
for $h=2$, $i \in \{1,2\}$, $h'=2$, $\tildeh \in \{2,4\}$, $\ti=4$, $\tildeh' \in \{5,7\}$.
There can be multiple such cases to consider and we can enforce all the PM constraints by setting the corresponding $\pi_{\tildeh \ti, h i}(\tildeh', h')$ variables equal to 0 as $\pi_{\tildeh \ti, h i}(\tildeh', h') = \bP(H_{hi} = h') \bP(H_{\tildeh \ti} = \tildeh' \mid H_{hi} = h')$.  As with CS (Proposition \ref{prop:CSBounds}),  PM will result in bounds $\PNUBPM$ and $\PNLBPM$ tighter than $\PNUB$ and $\PNLB$.

\begin{algorithm}
\begin{algorithmic}[1]
\REQUIRE{$(\fE, \fQ)$,   $(o_{1:T}, x_{1:T})$,  $B$, $\tx_{1:T}$}
\STATE{$h_{1:T}(b) \sim H_{1:T} \mid (o_{1:T}, x_{1:T}) \ \forall b = 1, \dots, B$}
\STATE{$\PNUB(B) = \max_{(\ftheta, \fpi) \in \cF} \text{PN}(\ftheta, \fpi \mid [h_{1:T}(b)]_b)$ }
\STATE{$\PNLB(B) = \min_{(\ftheta, \fpi) \in \cF} \text{PN}(\ftheta, \fpi \mid [h_{1:T}(b)]_b)$}
\STATE{\textbf{return} $(\PNLB(B), \PNUB(B))$}
\end{algorithmic}
\caption{Counterfactual analysis via optimization}
\label{alg:summary}
\end{algorithm}

We summarize our developments in Algorithm \ref{alg:summary}, which outputs the bounds $(\PNLB(B), \PNUB(B))$\footnote{We can output $(\PNLBCS(B), \PNUBCS(B))$ and $(\PNLBPM(B), \PNUBPM(B))$ as well by solving the same optimization problems but with additional linear constraints.}.
Line 1 (sampling) can be executed efficiently (cf.\ \S\ref{sec:sampling}), and we discuss three computational considerations behind solving the polynomial optimizations (lines 2 and 3).

First, though the constraints are linear, the objective is polynomial, making it a non-trivial non-convex optimization problem.  To solve it, we leverage state-of-the-art developments in optimization. In particular, we use the \texttt{BARON} solver \citep{baron}, which relies on a polyhedral branch-and-cut approach, allowing it to achieve global optima \citep{tawarmalani2005polyhedral}.  We found it to work well in our numeric experiments (\S\ref{sec:exp}).

Second,  in terms of the problem size, it follows from \eqref{eq:PairwiseMarginals} and \eqref{eq:JointVariables} that we have \emph{at most} $\lvert \bH \rvert^4 \lvert \bO \rvert^2 + \lvert \bH \rvert^2 \lvert \bO \rvert^2 \lvert \bX \rvert^2$ pairwise variables and $\lvert \bO \rvert^{\lvert \bH \rvert \lvert \bX \rvert} + \lvert \bH \rvert^{\lvert \bH \rvert \lvert \bO \rvert}$ joint variables.
Similarly, it follows from \eqref{eq:marginalconsthetapi} and \eqref{eq:pairwisemarginallink} that the feasible region $\cF$ is defined by \emph{at most} $\lvert \bO \rvert \lvert \bH \rvert \lvert \bX \rvert  + \lvert \bH \rvert^2 \lvert \bO \rvert + \lvert \bO \rvert^2 \lvert \bH \rvert^2 \lvert \bX \rvert^2 + \lvert \bH \rvert^4 \lvert \bO \rvert^2$ constraints.
However, these are merely upper bounds and we can exploit the sparsity inherent in the underlying application (along with the variable and constraint elimination discussed in Footnote \ref{ft:elimination}) to drastically reduce the problem size. For instance,  in our breast cancer application, we have $(\lvert \bH \rvert, \lvert \bO \rvert, \lvert \bX \rvert) = (7,7,2)$, with the above formulae giving over $10^{41}$ variables and $10^5$ constraints.  After we exploit sparsity (discussed in \S\ref{sec:exp}),  they are reduced to 16,124 and 610, respectively.
Further, as CS and PM can be modeled by setting appropriate variables to 0, they allow for further sparsity as we can delete those variables.

Third,  observe that a naive expansion of the recursion in Lemma \ref{lem:PNrecursive} results in a number of terms that is exponential in $T$, which would result in memory issues for moderate to large values of $T$. Nonetheless, as we elaborate in \S\ref{sec:reformulationApp}, it is possible to remove this exponential dependence on $T$ by a reformulation of the optimization problem. This comes at the cost of introducing polynomial constraints.  Nonetheless,  this reformulation allowed us to obtain high-quality solutions in the breast cancer setting with as many as $T = 100$ periods (\S\ref{sec:CompPerformance}). In contrast, we run into memory issues for $T$ as small as 11 with the original formulation.
In fact, we discuss an alternative approach at the end of \S\ref{sec:reformulationApp}. This approach allows us to compute the objective function efficiently without having to add any additional constraints. Unfortunately, the \texttt{BARON} solver does not allow us to use this approach and so we leave this issue for future research.

\section{Numerical Experiments}  \label{sec:exp}
We now apply our approach to the breast cancer application we described in \S\ref{sec:intro}.

\paragraph{Setup.} We described the elements of the underlying dynamic latent-state model $\fM \equiv (\fp, \fE, \fQ)$ in \S\ref{sec:problem}. It has a total of 7 states, 7 emissions, and 2 actions.
Given patient-level data $(o_{1:T}, x_{1:T})$, we wish to estimate the PN as defined in \eqref{eq:PN}.  The primitives $(\fp, \fE, \fQ)$ are calibrated to real-data using a mix of sources,  which we discuss in \S\ref{sec:CancerHMMPrimitives}.


\begin{figure*}
  \begin{center}
    \subfigure[UB / LB]{\label{fig:PN3a}\includegraphics[width=.3\linewidth]{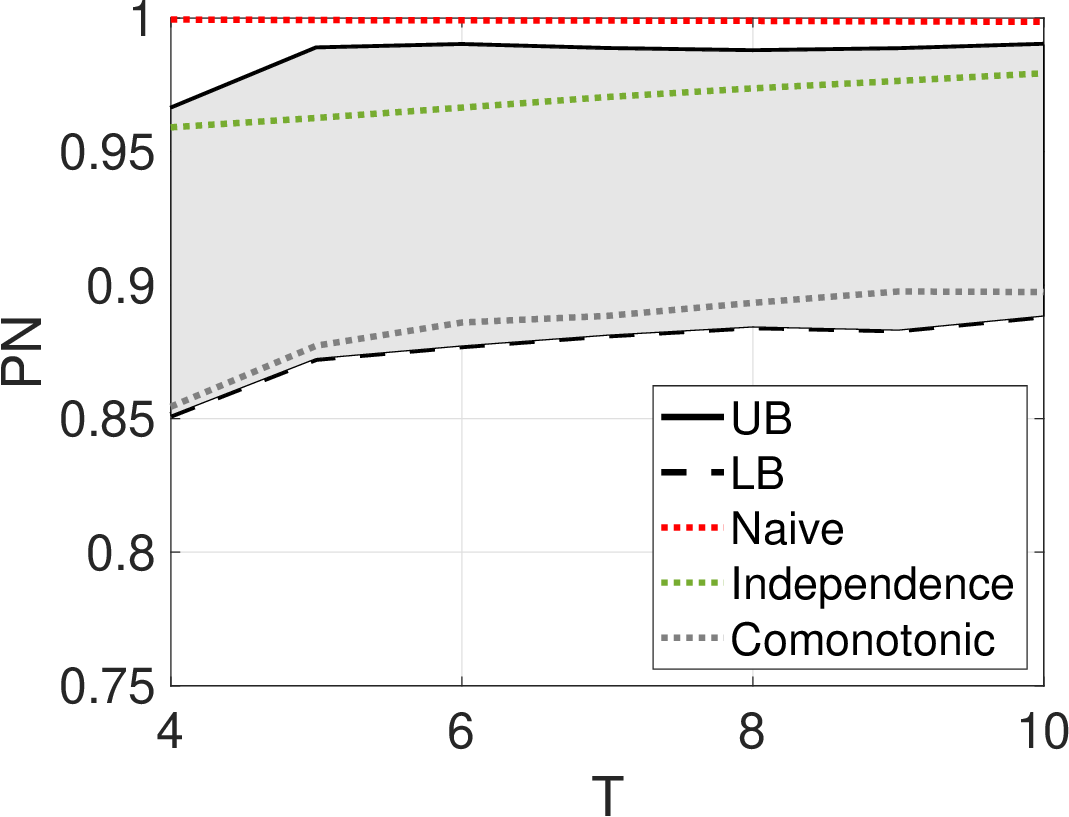}}
    \quad
    \subfigure[UB / LB with CS]{\label{fig:PN3b}\includegraphics[width=.3\linewidth]{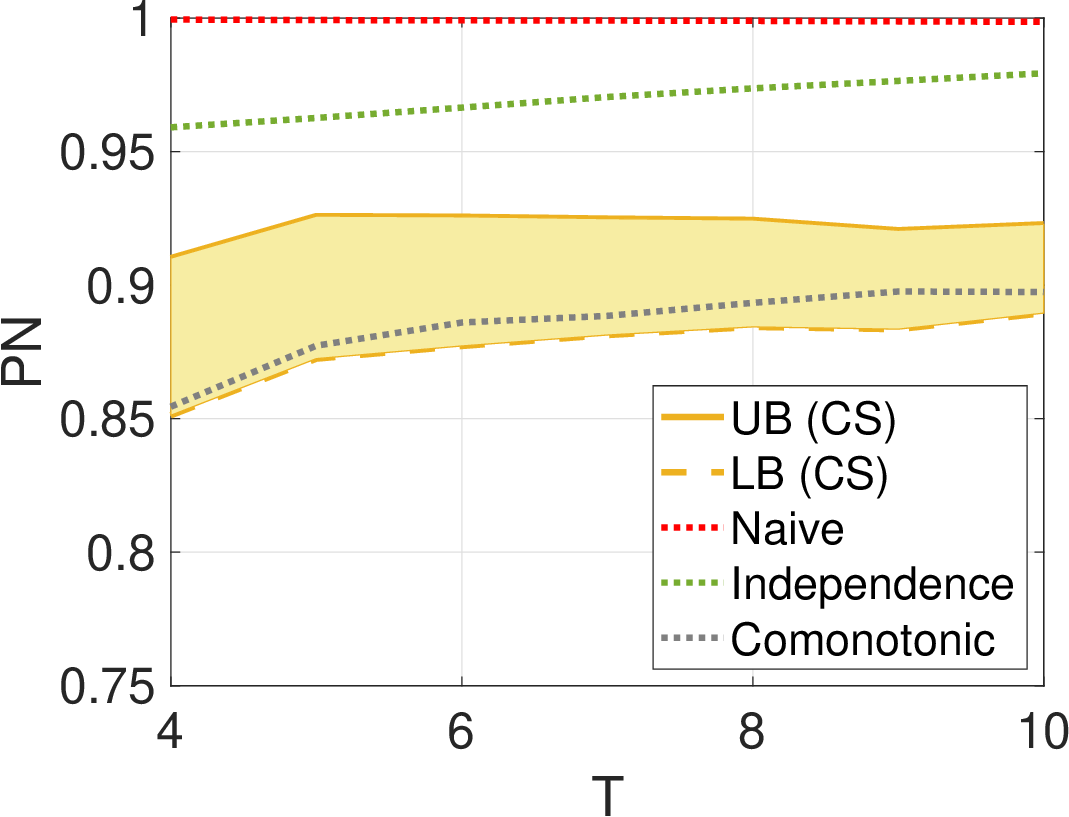}}
    \quad
    \subfigure[UB / LB with PM]{\label{fig:PN3c}\includegraphics[width=.3\linewidth]{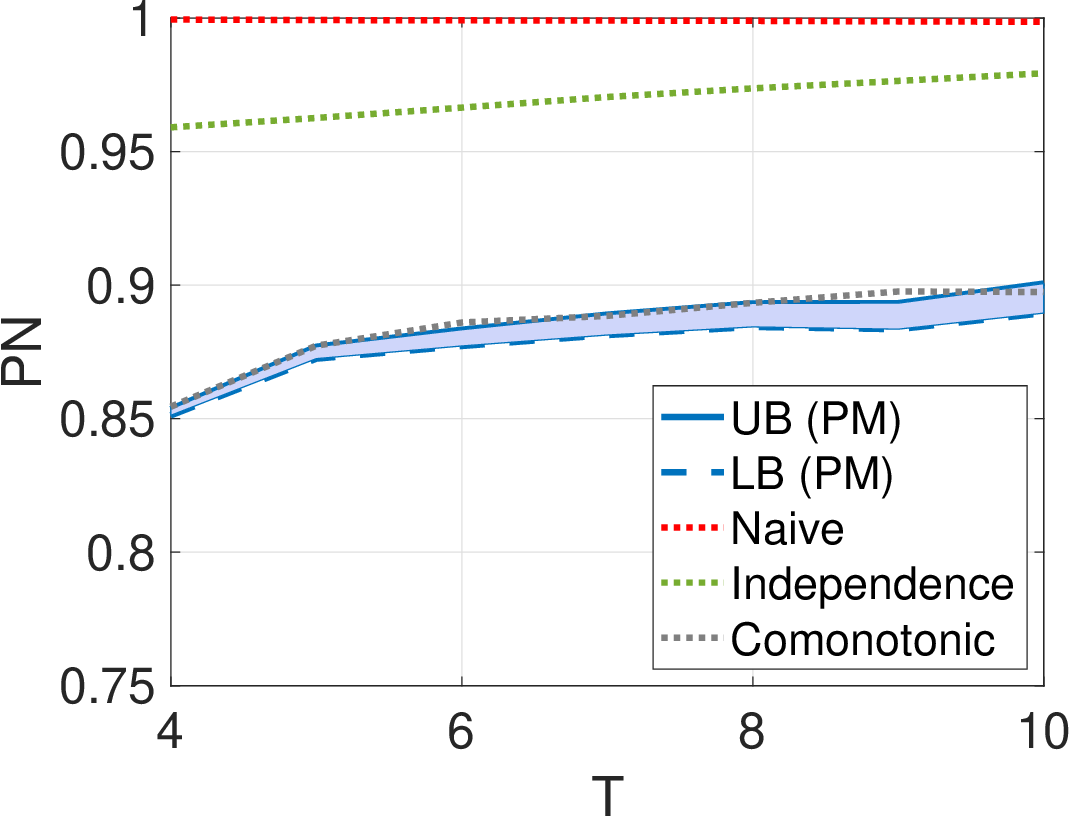}}
  \end{center}
\caption{PN results for path 1 as we vary $T \in \{4,\ldots,10\}$.
Observe that the \texttt{LB}, \texttt{LB(CS)}, and \texttt{LB(PM)} curves coincide (the lowest curve in each figure).
We show the average over 20 seeds and note that the standard deviation (s.d.) for every data point is smaller than 0.01.
Given this small magnitude, we omit the $\pm 1$ s.d. bars to reduce the clutter in the sub-figures.
Note that to simulate the \texttt{naive} estimate and the two copulas (\texttt{independence} and \texttt{comonotonic}), we use $10^4$ Monte-Carlo samples.  These simulations were fast (a couple of minutes).  Using $10^4$ samples is in contrast to the 100 samples we use for SAA and we did so to ensure a low s.d.
}
\label{fig:PN3}
\end{figure*}

We consider two paths with the first path defined as:
\begin{align*}
\underbrace{o_1}_{=2},  \red{\underbrace{o_2,\ldots,o_{T-2}, o_{T-1}}_{=1}},  \underbrace{o_T}_{=7} \\
\underbrace{x_1}_{=0},  \red{\underbrace{x_2,\ldots,x_{T-2}, x_{T-1}}_{=1}},  \underbrace{x_T}_{=0}.
\end{align*}
That is,  we observe a negative test result in period 1 after which screening was not performed for $T-2$ periods (red font). The patient died from breast cancer in period $T$.
Note that under this path,  given the calibrated primitives in \S\ref{sec:CancerHMMPrimitives}, it has to be the case that $h_{T-1} = 3$ (undiagnosed invasive) since a transition from state 2 (undiagnosed in-situ) to 7 is impossible. Further, the transition from 3 to 7 is not unlikely ($q_{317} \approx 0.15$).
The second path is similar but with one difference: screening was performed in period $T-1$ and  invasive cancer was detected:
\begin{align*}
\underbrace{o_1}_{=2},  \red{\underbrace{o_2,\ldots,o_{T-2}}_{=1}}, \underbrace{o_{T-1}}_{=5},  \underbrace{o_T}_{=7} \\
\underbrace{x_1}_{=0},  \red{\underbrace{x_2,\ldots,x_{T-2}}_{=1}}, \underbrace{x_{T-1},  x_T}_{=0}.
\end{align*}
Hence, in contrast with path 1,  the final transition from invasive to death was under treatment with probability $q_{357} \approx 0.01$ (\S\ref{sec:CancerHMMPrimitives}), which is much smaller than $q_{317}$ from above.
Given that this low probability transition did occur, this suggests the patient had an ``aggressive'' cancer in path 2.
As such, regardless of what the optimal $\ftheta$ and $\fpi$ are, the chances of survival on the counterfactual path would be low because of this ``aggressive'' nature of the  cancer.
This doesn't hold on path 1 as the cancer was less ``aggressive''.

We vary $T \in \{4,\ldots,10\}$,  with a larger value of $T$ suggesting the cancer may have progressed more slowly.
We compute PN bounds using our framework (Algorithm \ref{alg:summary}), which we implemented in \texttt{MATLAB} \citep{MATLAB}.
The feasibility set $\cF$ over $(\ftheta,\fpi)$ corresponds to \eqref{eq:marginalconsthetapi}, \eqref{eq:pairwisemarginallink},  and \eqref{eq:nonnegativityjoint}.
To solve the polynomial optimizations,  we use the \texttt{MATLAB-BARON} interface \citep{baron} with \texttt{CPLEX} \citep{cplex} as the ``LP / MIP solver''.  It solved each of our problem instances to global optimality within minutes / hours (depending on $T$), with an ``absolute termination tolerance'' of $0.01$ (on an \texttt{Intel Xeon E5} processor with 16 GB RAM).  Optimizations for $T=10$ took the longest time on average ($\sim$2 hours).
We generated $B=100$ samples using our sampling method in \S\ref{sec:sampling}. It took less than a second and we found $B=100$ was large enough to produce stable results for our SAA. We ensured this stability by computing our results for 20 seeds (for each (path, $T$) pair) and verifying the standard deviations to be small.
Though stability over the seeds is important, our PN estimates may still be biased for a finite $B$ (recall Proposition \ref{prop:SAA} only holds asymptotically).  As a check, we also generated results for $B=500$ and observed them to be very similar to the ones for $B=100$.
As noted below Algorithm \ref{alg:summary}, the sparse structure of $\fE$ and $\fQ$ drastically reduces the size of the problem.  For example,  when considering the $\pi_{\tildeh \ti, hi}(\tildeh', h')$ variables, we rule out the ones that map to impossible $(h,i,h')$ or $(\tildeh, \ti, \tildeh')$ combinations (refer to \S\ref{sec:CancerPrimitiveQ}). The same observation also applies to all the joint variables (details in \S\ref{sec:jointsparsity}).

\paragraph{Results.} The results for path 1 are displayed in Figure \ref{fig:PN3} (and for path 2 in Figure \ref{fig:PN4} (\S\ref{sec:pathtworesults})), where we show the PN bounds as we vary $T$. In addition to the bounds $(\PNLB, \PNUB)$ computed via our baseline optimization (\texttt{UB} and \texttt{LB}), we show the bounds obtained when we encode CS (\texttt{UB(CS)} and \texttt{LB(CS)}) and PM (\texttt{UB(PM)} and \texttt{LB(PM)})\footnote{Details on the PM constraints for breast cancer are in \S\ref{sec:PMCancer}.}. We also show the PN estimate when we perform counterfactual simulations using the two copulas discussed in \S\ref{sec:CopulaAppendix} (\texttt{independence} and \texttt{comonotonic}\footnote{Further details on the comonotonic copula specific to the breast cancer model are in \S\ref{sec:comonocancer}.}). Finally, the \texttt{naive} estimate  completely ignores the information in the observations, i.e., it does not execute the abduction step and is therefore an invalid estimate of PN.

To simplify matters, we adopt an all-or-nothing approach whereby either CS is imposed for both hidden-state transitions {\em and} observations or not at all. We do the same for PM. Of course, it is possible to consider various combinations, e.g., imposing PM for hidden-state transitions only or imposing CS only for the observations, etc. This is also true of our copulas when we estimate PN for a particular SCM. In Figure \ref{fig:PN3}, for example,
the \texttt{independence} (\texttt{comonotonic}) curve corresponds to assuming the independence (comonotonic) copula for both hidden-state transitions {\em and} observations. But we could of course have assumed one copula for the hidden-state transitions and an entirely different one for the observations. Each such combination of copulas would yield a different SCM and therefore a {\em feasible} value of PN.

The \texttt{naive} estimate is independent of the observed path and can fall outside the bounds. This makes sense as it does not perform abduction but simply simulates the original model $\fM$ under the intervention policy $\tx_{1:T} = 0$.  The naive estimates are very close to 1 as dying of breast cancer in any 5-year period\footnote{Each period maps to $6$ months so $T=10$ maps to $5$ years.} is highly unlikely.

For path 1,  we obtain relatively tight bounds, with PN always above 0.85.
This means that in the counterfactual world, the patient would have not died with high probability,  consistent with our discussion around $q_{317}$ above.
Even in the absence of any additional structure such as CS or PM, the gap between the lower and upper bounds is within $\sim$10 percentage points. The gap gets tighter with CS (within $\sim$5 percentage points) and PM (within $\sim$1 percentage point!).
The fact that the LB and UB under CS do not coincide resolves the open question of \citet{ICML209-Obers} regarding the uniqueness of the Gumbel-max mechanism w.r.t.\ CS -- it is not unique.
It is not surprising that the comonotonic estimate falls close to the PM bounds. Interestingly, the estimated PN for the two copulas roughly cover the range of possibilities in terms of the bounds (Figure \ref{fig:PN3a}).

For path 2,  the lower bounds are close to 0.
This aligns with the fact that despite being diagnosed in period $T-1$ (and hence, provided treatment), the patient eventually died (which suggests that the patient had an ``aggressive'' cancer).
The bounds without CS and PM are relatively loose, simply reflecting the lack of knowledge to reason in a counterfactual world.  As soon as we inject knowledge via CS or PM,  the bounds become much tighter.

The experiments discussed so far are for up to $T=10$ and we run into memory issues for $T > 10$ (recall the discussion at the end of \S\ref{sec:optProblem}).
Nonetheless, as we show in \S\ref{sec:scalability}, we can enhance the scalability of the polynomial optimizations in \eqref{eq:MaxMin} via a reformulation and an approximation. In fact,  as we demonstrate via numerics, these ideas allow us to obtain high-quality solutions for $T$ as large as $100$ in just a few hours of compute time.

\section{Concluding Remarks}  \label{sec:conc}
We have provided a framework for performing counterfactual analysis in dynamic latent-state models and in particular, computing lower and upper bounds on CQIs.
There are several interesting directions for future research.
First, we would like to handle the objective function in the optimization more efficiently as discussed at the end of \S\ref{sec:optProblem}. Specifically, \texttt{BARON}'s solver appears to explicitly expand the objective function which results in a number of terms that is exponential in $T$. We were able to finesse this issue in \S\ref{sec:scalability} via a reformulation but we suspect the approach outlined at the end of \S\ref{sec:optProblem} might provide a better solution.  All told, it may therefore be worthwhile developing an optimization algorithm specifically tailored to the problem (a polynomial objective with linear constraints) rather than using an off-the-shelf solver.
Another possible direction is exploring the use of variance reduction methods and other Monte-Carlo techniques to improve our basic Monte Carlo approach for generating posterior sample paths.
Finally, on the practical front, it would be of interest to apply our framework to real-world medical applications and use domain-specific knowledge to obtain (via the imposition of additional constraints) tighter bounds on the CQIs.

\section*{Acknowledgements}
We thank the ICML review team,  Madhumitha Shridharan, and Jim Smith for taking the time to read the paper and providing very useful feedback. We also thank Nick Sahinidis for his support with \texttt{BARON}-related issues.

\bibliography{bibliography}
\bibliographystyle{icml2023}

\newpage
\onecolumn
\appendix


\section{Proofs}  \label{sec:proofs}

\PNdecomp*
\textbf{Proof.} Observe that
\begin{subequations}
\begin{align*}
\text{PN} &= \Pb_{\tH_T}(\tH_T \neq 7) \tag*{[by definition]} \\
    &= 1 - \Pb_{\tH_T}(\tH_T = 7) \tag*{[$\bP(Y \neq y) = 1 - \bP(Y = y)$]}  \\
   &= 1 - \bE_{\tH_T}[\bI\{ \tH_T = 7 \}] \tag*{[$\bP(Y=y) = \bE[\bI\{Y=y\}]]$} \\
   &= 1 - \bE_{H_{1:T}} [\bE_{\tM \mid H_{1:T}} [\bI\{\tH_T = 7\} ]] \tag*{[law of total expectation]} \\
   &= 1 - \frac{1}{B} \sum_{b=1}^B \Pb_{\tM(b)} (\tH_T = 7)  \text{ as } B \to \infty. \tag*{[law of large numbers]}
\end{align*}
\end{subequations}
The proof is now complete. \hfill \qed

\leaveline

\PNrecursive*
\textbf{Proof.} For $t \in \{T, T-1, \ldots, 2\}$, observe that
\begin{subequations}
\begin{align*}
\Pb_{\tM(b)} (\tildeh_t) &=  \sum_{\tildeh_{t-1} \in \bH} \sum_{\tildeo_{t-1} \in \bO} \Pb_{\tM(b)} (\tildeh_t,  \tildeh_{t-1}, \tildeo_{t-1} ) \\
   &= \sum_{\tildeh_{t-1} \in \bH} \sum_{\tildeo_{t-1} \in \bO} \Pb_{\tM(b)}(\tildeh_t \mid \tildeh_{t-1}, \tildeo_{t-1}) \Pb_{\tM(b)} (\tildeo_{t-1} \mid \tildeh_{t-1}) \Pb_{\tM(b)} (\tildeh_{t-1} ) \\
    &= \sum_{\tildeh_{t-1} \in \bH} \sum_{\tildeo_{t-1} \in \bO} \tq^{(t-1)}_{\tildeh_{t-1} \tildeo_{t-1} \tildeh_t}(b) \times  \te^{(t-1)}_{\tildeh_{t-1} \tx_{t-1} \tildeo_{t-1}}(b) \times \Pb_{\tM(b)} ( \tildeh_{t-1} ) \\
    &= \sum_{\tildeh_{t-1} \in \bH} \sum_{\tildeo_{t-1} \in \bO} \frac{\pi_{\tildeh_{t-1} \tildeo_{t-1}, h_{t-1}(b) o_{t-1}}(\tildeh_t, h_{t}(b))}{q_{h_{t-1}(b) o_{t-1} h_{t}(b)}} \times  \frac{\theta_{\tildeh_{t-1} \tx_{t-1}, h_{t-1}(b) x_{t-1}}(\tildeo_{t-1},  o_{t-1})}{e_{h_{t-1}(b) x_{t-1} o_{t-1}}} \times \Pb_{\tM(b)} ( \tildeh_{t-1} ).
\end{align*}
\end{subequations}
The base case ($t=1$) holds since the counterfactual hidden state in period 1 equals the posterior sample $h_1(b)$ (recall from \S\ref{sec:CFModel}).
The proof is now complete. \hfill \qed

\section{Sampling Hidden Paths from the Posterior Distribution}  \label{sec:sampling}

In this section, we show how one can efficiently perform filtering, smoothing, and sampling for the dynamic latent-state model in Figure \ref{fig:model}. As our model is a generalization of an HMM, these algorithms are simple generalizations of the standard variants corresponding to an HMM \citep{barber2012bayesian}.

\paragraph{Filtering.} We first compute $\alpha(h_t) := \Pb(h_t,o_{1:t}, x_{1:t})$ which will yield the un-normalized filtered posterior distribution. We can then easily normalize it to compute $\Pb \left( h_t \mid o_{1:t}, x_{1:t} \right) \propto \alpha(h_t)$. We begin with $\alpha(h_1):= \Pb(o_1 \mid h_1, x_1) \Pb(h_1 \mid x_1) \Pb(x_1) = \Pb(o_1 \mid h_1, x_1) \Pb(h_1) \Pb(x_1)$.  For $t > 1$,  note that
\begin{align*}
\alpha(h_t) &= \sum_{h_{t-1}} \Pb\left( h_t,h_{t-1}, o_{1:t-1}, o_t, x_{1:t}\right) \\
&= \sum_{h_{t-1}}  \Pb\left(o_t \mid h_t,h_{t-1}, o_{1:t-1}, x_{1:t}\right)  \Pb\left( h_t \mid h_{t-1}, o_{1:t-1}, x_{1:t} \right)  \Pb(x_t \mid h_{t-1}, o_{1:t-1}, x_{1:t-1})  \Pb\left( h_{t-1}, o_{1:t-1}, x_{1:t-1}\right)  \\
&= \sum_{h_{t-1}}  \Pb\left(o_t \mid h_t, x_t \right)  \Pb\left( h_t \mid h_{t-1}, o_{t-1} \right)  \Pb(x_t)  \Pb\left( h_{t-1}, o_{1:t-1}, x_{1:t-1}\right)  \\
&= \Pb(x_t)  \Pb\left(o_t \mid h_t, x_t \right)  \sum_{h_{t-1}}  \Pb\left( h_t \mid h_{t-1}, o_{t-1} \right)  \alpha(h_{t-1}).
\end{align*}

\paragraph{Smoothing.} We now compute $\beta(h_t):= \Pb(o_{t+1:T}, x_{t+1:T} \mid h_t,o_t)$ with the understanding that $\beta(h_T) = 1$.  For $t < T$, we have
\begin{align*}
\beta(h_t) &= \sum_{h_{t+1}} \Pb(o_{t+1},x_{t+1},o_{t+2:T}, x_{t+2:T}, h_{t+1} \mid h_t, o_t)  \\
&= \sum_{h_{t+1}} \Pb(o_{t+2:T}, x_{t+2:T}  \mid h_t, o_t, h_{t+1},o_{t+1},x_{t+1})  \Pb(h_{t+1},o_{t+1},x_{t+1}\mid h_t, o_t)  \\
&= \sum_{h_{t+1}} \Pb(o_{t+2:T}, x_{t+2:T}  \mid h_{t+1},o_{t+1})  \Pb(o_{t+1}\mid h_{t+1},x_{t+1}, h_t, o_t)  \Pb(h_{t+1},x_{t+1}\mid h_t, o_t)  \\
&= \sum_{h_{t+1}} \beta(h_{t+1})  \Pb(o_{t+1}\mid h_{t+1},x_{t+1})  \Pb(x_{t+1}\mid h_{t+1},h_t, o_t)  \Pb(h_{t+1}\mid h_t, o_t)  \\
&= \Pb(x_{t+1})  \sum_{h_{t+1}} \beta(h_{t+1})  \Pb(o_{t+1}\mid h_{t+1},x_{t+1})   \Pb(h_{t+1}\mid h_t, o_t).
\end{align*}
Now, note that
\begin{align*}
\Pb\left(h_t, o_{1:T}, x_{1:T} \right) &= \Pb\left(h_t, o_{1:t} , x_{1:t} \right)  \Pb\left(  o_{t+1:T}, x_{t+1:T} \mid h_t, o_{1:t}, x_{1:t}\right)  \\
&= \Pb\left(h_t, o_{1:t}, x_{1:t} \right)  \Pb\left( o_{t+1:T},x_{t+1:T} \mid h_t, o_t\right)  \\
&= \alpha(h_t) \beta(h_t).
\end{align*}
We therefore obtain the hidden state marginal
\begin{align*}
\Pb\left(h_t \mid o_{1:T}, x_{1:T} \right)  =  \frac{\alpha(h_t) \beta(h_t)}{\sum_{h_t} \alpha(h_t) \beta(h_t)},
\end{align*}
which solves the smoothing problem.

\paragraph{Pairwise marginal.}
We can compute $\Pb\left(h_t,h_{t+1} \mid o_{1:T},x_{1:T} \right)$ by noting that
\begin{align}
\Pb\left(h_t,h_{t+1} \mid o_{1:T},x_{1:T} \right) & \propto   \Pb\left(o_{1:t}, o_{t+1}, o_{t+2:T}, x_{1:t}, x_{t+1}, x_{t+2:T}, h_{t+1}, h_t \right) \nonumber \\
&= \Pb\left(o_{t+2:T},x_{t+2:T} \mid o_{1:t}, o_{t+1}, x_{1:t}, x_{t+1}, h_{t+1}, h_t \right)  \Pb\left( o_{1:t}, o_{t+1},x_{1:t}, x_{t+1},  h_{t+1}, h_t \right) \nonumber \\
&= \Pb\left(o_{t+2:T},x_{t+2:T} \mid  h_{t+1},o_{t+1} \right)  \Pb\left(o_{t+1} \mid o_{1:t},  h_{t+1}, h_t,x_{1:t}, x_{t+1} \right)  \Pb\left(o_{1:t},  h_{t+1}, h_t,x_{1:t}, x_{t+1} \right) \nonumber \\
&= \Pb\left(o_{t+2:T},x_{t+2:T} \mid  h_{t+1},o_{t+1} \right)  \Pb\left(o_{t+1} \mid h_{t+1},x_{t+1} \right)  \Pb\left(h_{t+1},x_{t+1} \mid o_{1:t},x_{1:t},  h_t \right)  \Pb\left(o_{1:t},x_{1:t},  h_t \right) \nonumber \\
&= \Pb\left(o_{t+2:T},x_{t+2:T} \mid  h_{t+1},o_{t+1} \right)  \Pb\left(o_{t+1} \mid h_{t+1},x_{t+1} \right)  \Pb\left(h_{t+1}, x_{t+1} \mid  h_t,o_t \right)  \Pb\left(o_{1:t}, x_{1:t}, h_t \right). \label{eq:PairwiseMarginal}
\end{align}
We can rearrange (\ref{eq:PairwiseMarginal}) to obtain
\begin{align}
\Pb\left(h_t,h_{t+1} \mid o_{1:T}, x_{1:T} \right)  \propto  \alpha(h_t) \Pb\left(o_{t+1} \mid h_{t+1},x_{t+1} \right) \Pb\left(h_{t+1},x_{t+1} \mid  h_t,o_t \right) \beta(h_{t+1}).  \label{eq:PairwiseMarginal1}
\end{align}
Therefore, $\Pb\left(h_t,h_{t+1} \mid o_{1:T}, x_{1:T} \right)$ is easy to compute once the forward-backward, i.e. the filtering and smoothing, recursions have been completed.

\paragraph{Sampling.} We would like to sample from the posterior $\Pb\left( h_{1:T} \mid o_{1:T}, x_{1:T} \right)$. We can do this by first noting that
\begin{align*}
\Pb\left( h_{1:T} \mid o_{1:T}, x_{1:T} \right) &= \Pb\left( h_1 \mid h_{2:T}, o_{1:T}, x_{1:T} \right)  \ldots  \Pb\left( h_{T-1} \mid h_T, o_{1:T}, x_{1:T} \right) \Pb\left( h_T \mid o_{1:T}, x_{1:T} \right)  \\
&= \Pb\left( h_1 \mid h_2, o_{1:T}, x_{1:T} \right) \ldots \Pb\left( h_{T-1} \mid h_T, o_{1:T}, x_{1:T} \right) \Pb\left( h_T \mid o_{1:T}, x_{1:T} \right).
\end{align*}

We can therefore sample sequentially via the following two steps:
\begin{itemize}
\item First, draw $h_T$ from $\Pb\left( h_T \mid o_{1:T}, x_{1:T} \right)$, which we know from the smoothed distribution of $h_T$.
\item Second, observe that for any $t < T$, we have
\begin{align*}
\Pb\left( h_t \mid h_{t+1}, o_{1:T}, x_{1:T} \right) & \propto \Pb\left( h_t , h_{t+1} \mid o_{1:T}, x_{1:T} \right) \\
& \propto \alpha(h_t) \Pb\left(h_{t+1},x_{t+1} \mid  h_t,o_t \right)  \tag*{[by \eqref{eq:PairwiseMarginal1}]} \\
&= \alpha(h_t) \Pb\left(h_{t+1} \mid  h_t,o_t \right)  \Pb\left(x_{t+1}\right),
\end{align*}
from which it is easy to sample.
\end{itemize}
Hence, we can efficiently generate samples $[h_{1:T}(b)]_{b=1}^B$ from the posterior $\Pb\left( h_{1:T} \mid o_{1:T}, x_{1:T} \right)$.

\section{A Brief Introduction to Copulas and Counterfactual Simulations}  \label{sec:CopulaAppendix}

Copulas are functions that enable us to separate the marginal distributions from the dependency structure of a given multivariate distribution. They are particularly useful in applications where the marginal distributions are known (either from domain specific knowledge or because there is sufficient marginal data) but a joint distribution with these known marginals is required.
In our application in this paper, we know the marginal distribution of each random variable in $[O_{hx}]_{h,x}$ and $[H_{hi}]_{h,i}$, which is dictated by the model primitives $(\fE, \fQ)$ as follows: $e_{hxi} = \bP(O_{hx} = i)$ and $q_{hih'} = \bP(H_{hi} = h')$.
Indeed, these marginal distributions can be estimated from data, but the {\em joint distribution} must be specified in order to compute counterfactuals.

In each of these cases, one needs to work with a joint distribution with fixed or pre-specified marginal distributions. Copulas and Sklar's Theorem (see below) can be very helpful in these situations. We only briefly review some of the main results from the theory of copulas here but \citet{Nelsen2006} can be consulted for an introduction to the topic. \citet{QRM2015} also contains a nice introduction but in the context of financial risk management.

\begin{defi} \label{def:Copula}
A $d$-dimensional {copula}, $C : [0,1]^d : \rightarrow [0,1]$ is a cumulative distribution function with uniform marginals.
\end{defi}

We write $C({\bf u}) = C(u_1, \ldots , u_d)$ for a generic copula. It follows immediately from Definition \ref{def:Copula} that  $C(u_1, \ldots , u_d)$ is non-decreasing in each argument and that $C(1,\ldots , 1, u_i, 1, \ldots , 1)  =  u_i$. It is also easy to confirm that $C(1,u_1, \ldots , u_{d-1})$ is a $(d-1)$-dimensional copula and, more generally, that all $k$-dimensional marginals with $2 \leq k \leq d$ are copulas.  The most important result from the theory of copulas is Sklar's Theorem \cite{Sklar1959}.

\begin{thm}[{\bf Sklar 1959}]
Consider a $d$-dimensional CDF $\bPi$ with marginals $\bPi_1$, \ldots , $\bPi_d$. Then, there exists a copula $C$ such that
\begin{equation}
\bPi(x_1, \ldots , x_d) = C\left(\bPi_1(x_1), \ldots , \bPi_d(x_d)  \right)  \label{eq:Sklar1}
\end{equation}
for all $x_i \in [-\infty, \; \infty]$ and $i=1, \ldots, d.$

If $\bPi_i$ is continuous for all $i = 1, \ldots , d$, then $C$ is unique; otherwise $C$ is uniquely determined only on $\mbox{Ran}(\bPi_1) \times \cdots \times \mbox{Ran}(\bPi_d)$, where $\mbox{Ran}(\bPi_i)$ denotes the range of the CDF $\bPi_i$.

Conversely, consider a copula $C$ and univariate CDF's $\bPi_1, \ldots , \bPi_d$. Then, $\bPi$ as defined in (\ref{eq:Sklar1}) is a multivariate CDF with marginals $\bPi_1, \ldots , \bPi_d$.
\end{thm}

A particularly important aspect of Sklar's Theorem in the context of this paper is that $C$ is only uniquely determined on $\mbox{Ran}(\bPi_1) \times \cdots \times \mbox{Ran}(\bPi_d)$. Because we are interested in applications with discrete state-spaces, this implies that there will be many copulas that lead to the same joint distribution $\bPi$. It is for this reason that we prefer to work directly with the joint distribution of $[O_{hx}]_{h,x}$ and $[H_{hi}]_{h,i}$ (recall \eqref{eq:PairwiseMarginals}). That said, we emphasize that specifying copulas for the exogenous vectors $\fU_t$ and $\fV_t$ is equivalent to specifying a particular structural causal model (SCM) in which any CQI can be computed.

The following important result was derived independently by Fr\'{e}chet and Hoeffding and provides lower and upper bounds on copulas.
\begin{thm}[{\bf The Fr\'{e}chet-Hoeffding Bounds}] Consider a copula $C({\bf u}) = C(u_1, \ldots , u_d)$. Then,
\begin{align*}
\max \left\{1 - d + \sum_{i=1}^d u_i,  0  \right \}  \leq  C({\bf u})  \leq \ \min \{u_1, \ldots , u_d \}.
\end{align*}
\end{thm}

Three important copulas are the comonotonic, countermonotonic (only when $d=2$) and independence copulas which model extreme positive dependency, extreme negative dependency and (not surprisingly) independence. They are defined as follows.

\paragraph{Comonotonic Copula.} The  comonotonic copula is given by
\begin{equation}  \label{eq:CoM1}
\CP({\bf u}) := \min \{u_1, \ldots , u_d  \},
\end{equation}
which coincides with the Fr\'{e}chet-Hoeffding upper bound. It corresponds to the case of  extreme positive dependence. For example, let $\fU =(U_1, \ldots , U_d)$ with $U_1 = U_2 = \cdots = U_d  \sim \text{Unif}[0,1]$. Then, clearly $\min \{u_1, \ldots , u_d  \} = \bPi(u_1,\ldots , u_d)$ but by Sklar's Theorem $F(u_1,\ldots , u_d) = C(u_1,\ldots , u_d)$ and so, $ C(u_1,\ldots , u_d) = \min \{u_1, \ldots , u_d  \}$.

\paragraph{Countermonotonic Copula.} The  countermonotonic copula is a 2-dimensional copula given by
\begin{align}  \label{eq:CounterM1}
\CN({\bf u}) := \max \{u_1 + u_2 - 1,   0 \},
\end{align}
which coincides with the Fr\'{e}chet-Hoeffding lower bound when $d=2$. It corresponds to the case of extreme negative dependence.  It is easy to check that (\ref{eq:CounterM1}) is the joint distribution of $(U,1-U)$ where $U \sim \text{Unif}[0,1]$. (The Fr\'{e}chet-Hoeffding lower bound is only tight when $d=2$. This is analogous to the fact that while a pairwise correlation can lie anywhere in $[-1,1]$, the {\em average} pairwise correlation of $d$ random variables is bounded below by $-1/(d-1)$.)

\paragraph{Independence Copula.}  The independence copula satisfies
\begin{align*}
\CI({\bf u})  :=  \prod_{i=1}^d u_i,
\end{align*}
and it is easy to confirm using Sklar's Theorem that random variables are independent if and only if their copula is the independence copula.

\leaveline

A well known and important result regarding copulas is that they are invariant under monotonic transformations.
\begin{prop}[{\bf Invariance Under Monotonic Transformations}] Suppose the random variables $X_1, \ldots , X_d$ have continuous marginals and  copula $C_X$. Let $T_i  :  \mathbb{R} \rightarrow \mathbb{R}$, for $i=1, \ldots , d$ be strictly increasing functions.
Then, the dependence structure of the random variables
\begin{align*}
Y_1:=T_1(X_1),  \ldots ,  Y_d:=T_d(X_d)
\end{align*}
is also given by the copula $C_X$.
\end{prop}

This leads immediately to the following result.

\begin{prop} Let $X_1, \ldots , X_d$ be random variables with continuous marginals and suppose $X_i = T_i(X_{{1}})$ for $i=2, \ldots , d$ where $T_2, \ldots , T_d$ are strictly increasing transformations. Then, $X_1, \ldots , X_d$ have the comonotonic copula.
\end{prop}
{\bf Proof.} Apply the {\em invariance under monotonic transformations} proposition and observe that the copula of $(X_1, X_1, \ldots , X_1)$ is the comonotonic copula. \hfill \qed

Our optimization framework implicitly optimizes over the space of copulas by solving polynomial programs with possibly a large  number of variables and constraints. (We saw in \S\ref{sec:optProblem} that the number of variables and constraints is polynomial in $\lvert \bH \rvert$, $\lvert \bO \rvert$ and $\lvert \bX \rvert$ when calculating the probability of necessity (PN).) It may also be worthwhile, however, working explicitly with copulas. For example, the independence and comonotonic copulas are well understood and using these copulas to define SCMs may provide interesting benchmarks. Indeed, we estimate the PN for these benchmarks in our numerical results of \S\ref{sec:exp}.
Towards this end, in \S\ref{sec:CopulaSimInd} and \S\ref{sec:CopulaSimComon}, we explain how we can simulate our dynamic latent-state model to estimate the CQI under the independence (\S\ref{sec:CopulaSimInd}) and comonotonic (\S\ref{sec:CopulaSimComon}) copulas. Specifically, we assume each of the copulas for $\fU_t$ and $\fV_t$ are the independence copulas in \S\ref{sec:CopulaSimInd},  whereas in \S\ref{sec:CopulaSimComon}, we assume their copulas are the comonotonic copula.

There is no reason, however, why we couldn't combine them and assume, for example, that the copula for $\fU_t$ was the independence copula and the copula for $\fV_t$ was the comonotonic copula. More generally, we could use domain-specific knowledge to identify or narrow down sub-components of the copulas and leave the remaining components to be identified  via the optimization problems. Since convex combinations of copulas are copulas, we could also optimize over such combinations. For example, suppose domain specific knowledge\footnote{It may be more likely that we only have domain specific knowledge over sub-components of the copulas (which are themselves copulas).} tells us that the copula of $\fV_t$ is $\lambda \CN + (1-\lambda) \CI$, i.e., a convex combination of the comonotonic and independence copulas, with $\lambda \in [0,1]$ unknown. Then, the optimization over $\fV_t$ would reduce to a single-variable ($\lambda$) optimization with a linear constraint. Of course, the optimization over the copula of $\fU_t$ must also be included but domain-specific knowledge may also help to simplify and constrain that component of the optimization. Properties such as pathwise monotonicity (PM) and counterfactual stability (CS) can also be expressed in copula terms. Indeed, PM can be expressed via the comonotonic copula, as we discuss in \S\ref{sec:CopulaSimComon}.

\subsection{Counterfactual Simulations Under the Independence Copula}  \label{sec:CopulaSimInd}

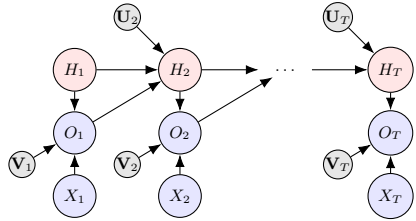
\begin{figure}[ht]
\centering
\begin{tikzpicture}[scale=.7,transform shape]
    \node[state,fill=red!10!white] (h1) at (0,0) {\footnotesize $H_1$};
    \node[state,fill=red!10!white] (h2) at (2,0) {\footnotesize $H_2$};
    \node[state2,fill=black!10!white] (u2) at (1,1) {\footnotesize $\fU_2$};
    \node[draw=none, minimum width = 1 cm] (hdots) at (4,0) {\footnotesize $\ldots$};
    \node[state,fill=red!10!white] (hT) at (6,0) {\footnotesize $H_T$};
    \node[state2,fill=black!10!white] (uT) at (5,1) {\footnotesize $\fU_T$};
    \node[state,fill=blue!10!white] (o1) at (0,-1.2) {\footnotesize $O_1$};
    \node[state2,fill=black!10!white] (v1) at (-1,-1.8) {\footnotesize $\fV_1$};
    \node[state,fill=blue!10!white] (o2) at (2,-1.2) {\footnotesize $O_2$};
    \node[state2,fill=black!10!white] (v2) at (1,-1.8) {\footnotesize $\fV_2$};
    \node[state,fill=blue!10!white] (oT) at (6,-1.2) {\footnotesize $O_T$};
    \node[state2,fill=black!10!white] (vT) at (5,-1.8) {\footnotesize $\fV_T$};
    \node[state,fill=blue!10!white] (x1) at (0,-2.4) {\footnotesize $X_1$};
    \node[state,fill=blue!10!white] (x2) at (2,-2.4) {\footnotesize $X_2$};
    \node[state,fill=blue!10!white] (xT) at (6,-2.4) {\footnotesize $X_T$};
    \path (h1) edge (o1);
    \path (h1) edge (h2);
    \path (o1) edge (h2);
    \path (h2) edge (o2);
    \path (h2) edge (hdots);
    \path (o2) edge (hdots);
    \path (hT) edge (oT);
    \path (hdots) edge (hT);
    \path (x1) edge (o1);
    \path (x2) edge (o2);
    \path (xT) edge (oT);
    \path (u2) edge (h2);
    \path (uT) edge (hT);
    \path (v1) edge (o1);
    \path (v2) edge (o2);
    \path (vT) edge (oT);
\end{tikzpicture}
\caption{The SCM underlying the dynamic latent-state model. }
\label{fig:SCMApp}
\end{figure}

For convenience, we copy Figure \ref{fig:SCM} from the main text, which is now labelled as Figure \ref{fig:SCMApp}.  Furthermore, recall that $(o_{1:T}, x_{1:T})$ is the observed data and $\tx_{1:T}$ is the intervention policy that was applied.

As in \S\ref{sec:opt}, we start with the posterior samples $[h_{1:T}(b)]_{b=1}^B$ corresponding to the random path $H_{1:T} \mid (o_{1:T},  x_{1:T})$. These samples can be generated efficiently (cf.\ \S\ref{sec:sampling}).
For each sample $b$,  our goal is to convert the sampled path $h_{1:T}(b)$ into a counterfactual path $\tildeh_{1:T}(b)$.
As noted in \S\ref{sec:CFModel}, irrespective of the copula choice, the counterfactual hidden state in period 1 equals the posterior sample, i.e.,
\begin{align*}
\tildeh_1(b) = h_1(b).
\end{align*}
We next need to sample $\tildeh_2(b)$, but that first requires us to sample the counterfactual emission $\tildeo_1(b)$ (cf.\ Figure \ref{fig:SCMApp}). With the copula underlying $\fV_1$ being the independence copula, it follows that
\begin{align*}
\tildeo_1(b) = \begin{cases}
o_1 & \text{ if $x_1 = \tx_1$ and $h_1(b) = \tildeh_1(b)$} \\
\text{sample from the emission distribution $[e_{\tildeh_1(b) \tx_1 i}]_i$} & \text{ otherwise}.
\end{cases}
\end{align*}
The counterfactual emission $\tildeo_1(b)$ allows us to sample the counterfactual state $\tildeh_2(b)$, which again leverages the fact that the copula underlying $\fU_2$ is the independence copula:
\begin{align*}
\tildeh_2(b) = \begin{cases}
h_2(b) & \text{ if $h_1(b) = \tildeh_1(b) $ and $o_1 = \tildeo_1(b)$} \\
\text{sample from the transition distribution $[q_{\tildeh_1(b) \tildeo_1(b) h'}]_{h'}$} & \text{ otherwise.}
\end{cases}
\end{align*}
We then generate period 2 counterfactual emission $\tildeo_2(b)$ in a similar manner and the process repeats until we hit the end of horizon.  We summarize the procedure in Algorithm \ref{alg:CopulaSimInd}.

\begin{algorithm}[H]
\begin{algorithmic}[1]
\REQUIRE{$(\fE, \fQ)$,   $(o_{1:T}, x_{1:T})$,  $[h_{1:T}(b)]_{b=1}^B$, $\tx_{1:T}$}
\FOR{$b=1$ to $B$}
\STATE{$\tildeh_1(b) = h_1(b)$}
\FOR{$t=1$ to $T-1$}
\IF{$x_t = \tx_t$ and $h_t(b) = \tildeh_t(b)$}
\STATE{$\tildeo_t(b) = o_t$}
\ELSE
\STATE{$\tildeo_t(b) \sim \text{Categorical}([e_{\tildeh_t(b) \tx_t i}]_i)$}
\ENDIF
\IF{$h_t(b) = \tildeh_t(b) $ and $o_t = \tildeo_t(b)$}
\STATE{$\tildeh_{t+1}(b)  = h_{t+1}(b)$}
\ELSE
\STATE{$\tildeh_{t+1}(b)  \sim \text{Categorical}([q_{\tildeh_t(b) \tildeo_t(b) h'}]_{h'}$)}
\ENDIF
\ENDFOR
\ENDFOR
\STATE{\textbf{return} $[\tildeh_{1:T}(b)]_b$}
\end{algorithmic}
\caption{Counterfactual simulations under the independence copula}
\label{alg:CopulaSimInd}
\end{algorithm}

\subsection{Counterfactual Simulations Under the Comonotonic Copula}  \label{sec:CopulaSimComon}

\begin{figure}[ht]
\centering
\begin{tikzpicture}[scale=.7,transform shape]
    \node[state,fill=blue!10!white] (X) at (0,0) {\footnotesize $X$};
    \node[state,fill=blue!10!white] (Y) at (2,0) {\footnotesize $Y$};
    \node[state,fill=black!10!white] (U) at (1,1) {\footnotesize $U$};
    \path (X) edge (Y);
    \path (U) edge (Y);
\end{tikzpicture}
\caption{SCM for Example \ref{ex:CS} with the comonotonic copula and hence, the noise node is a scalar $U \sim \text{Unif}[0,1]$, as opposed to a vector $\fU$. The structural equation is $Y = f(X, U)$, which we denote by $f_X(U)$, the \emph{inverse transform} function corresponding to the random variable $Y_X$.  That is, $f_0(u) = $ bad, better,  and best if $u \in [0, 0.2]$,  $u \in [0.2, 0.5]$, and $u \in [0.5, 1]$, respectively.  Similarly, $f_1(u) = $ bad, better,  and best if $u \in [0, 0.2]$,  $u \in [0.2, 0.4]$, and $u \in [0.4, 1]$, respectively.}
\label{fig:SimpleExample}
\end{figure}
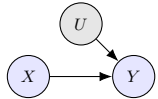

Before the formal description (which involves non-trivial notation), we provide the intuition (which is relatively straightforward).  We do so by revisiting Example \ref{ex:CS}, where we have the causal graph $X \to Y$ with $X \in \{0,1\}$ (medical treatment) and $Y \in \{\text{bad}, \text{better}, \text{best}\}$ (patient outcome). The outcome $Y_x := Y \mid (X = x)$ obeys the following distribution: $Y_0 \sim \{\text{bad}, \text{better}, \text{best}\}$ w.p.\ $\{0.2, 0.3, 0.5\}$ and $Y_1 \sim \{\text{bad}, \text{better}, \text{best}\}$ w.p.\ $\{0.2, 0.2, 0.6\}$.
The underlying SCM is shown again in Figure \ref{fig:SimpleExample}.

Consider a patient whose outcome $Y$ was ``better'' under no treatment ($x=0$).  Given the prior $U \sim \text{Unif}[0,1]$,  we get the posterior $U \mid (Y_0 = \text{better}) \sim \text{Unif}[0.2, 0.5]$. Now, suppose we are interested in the understanding the counterfactual outcome under the intervention $\tx = 1$, i.e., the random variable $\tY := Y_1 \mid (Y_0 = \text{better})$.  Then, given the $\text{Unif}[0.2, 0.5]$ belief over $U$ and the functional form of $f_1(\cdot)$ (as defined in the caption of Figure \ref{fig:SimpleExample}), we get that the $[0.2, 0.4]$ region of $U$ will map to ``better'' and the $[0.4, 0.5]$ to ``best''.
Hence,  $\tY$ equals ``better'' w.p.\ $2/3$ and ``best'' w.p.\ $1/3$.
This clearly obeys the pathwise monotonicity (PM) intuition we alluded to towards the end of Example \ref{ex:CS} (``the counterfactual outcome $\tY$ should not be worse under treatment ($\tx=1$) than under no treatment ($x=0$)'').

We now formalize this intuition to our dynamic latent-state model.
As a prerequisite to discussing the notion of PM, one needs to define an ordering of the states (set $\bH$) and the emissions (set $\bO$), e.g., from ``best'' to ``worst''. Denote by $r_{H}(h)$ the rank of state $h$ with respect to this ordering and by $r_{O}(i)$ the rank of emission $i$.
Furthermore, let $r_H^{-1}(r)$ and $r_O^{-1}(r)$ denote the inverse functions corresponding to $r_{H}(h)$ and $r_{O}(i)$, respectively. That is,  $r_H^{-1}(r)$ returns the state with rank $r$ and $r_O^{-1}(r)$ returns the emission with rank $r$.
Also, for each $(h,i)$ pair,  observe that $[q_{hih'}]_{h'}$ denotes the transition distribution (which maps to the random variable $H_{hi}$).  Corresponding to this distribution, define the rank-ordered CDF as follows:
\begin{subequations}
\begin{align}
Q_{hih'} := \sum_{h'' : r_H(h'') \le r_H(h')} q_{hih''} \ \forall h'.
\label{eq:RankCDFQ}
\end{align}
Similarly,  for each $(h,x)$ pair,  observe that $[e_{hxi}]_i$ denotes the emission distribution (which maps to the random variable $O_{hx}$).  Corresponding to this distribution, define the rank-ordered CDF as follows:
\begin{align}
E_{hxi} := \sum_{j : r_O(j) \le r_O(i)} e_{hxj} \ \forall i.
\label{eq:RankCDFE}
\end{align}
\end{subequations}
Also, define $Q_{hi0} = E_{hx0} = 0$ for all $(h,i)$ and $(h,x)$.
We discuss these orderings for the breast cancer application in \S\ref{sec:comonocancer}.

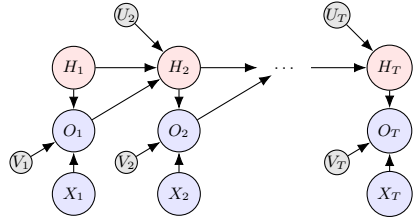
\begin{figure}[H]
\centering
\begin{tikzpicture}[scale=.7,transform shape]
    \node[state,fill=red!10!white] (h1) at (0,0) {\footnotesize $H_1$};
    \node[state,fill=red!10!white] (h2) at (2,0) {\footnotesize $H_2$};
    \node[state2,fill=black!10!white] (u2) at (1,1) {\footnotesize $U_2$};
    \node[draw=none, minimum width = 1 cm] (hdots) at (4,0) {\footnotesize $\ldots$};
    \node[state,fill=red!10!white] (hT) at (6,0) {\footnotesize $H_T$};
    \node[state2,fill=black!10!white] (uT) at (5,1) {\footnotesize $U_T$};
    \node[state,fill=blue!10!white] (o1) at (0,-1.2) {\footnotesize $O_1$};
    \node[state2,fill=black!10!white] (v1) at (-1,-1.8) {\footnotesize $V_1$};
    \node[state,fill=blue!10!white] (o2) at (2,-1.2) {\footnotesize $O_2$};
    \node[state2,fill=black!10!white] (v2) at (1,-1.8) {\footnotesize $V_2$};
    \node[state,fill=blue!10!white] (oT) at (6,-1.2) {\footnotesize $O_T$};
    \node[state2,fill=black!10!white] (vT) at (5,-1.8) {\footnotesize $V_T$};
    \node[state,fill=blue!10!white] (x1) at (0,-2.4) {\footnotesize $X_1$};
    \node[state,fill=blue!10!white] (x2) at (2,-2.4) {\footnotesize $X_2$};
    \node[state,fill=blue!10!white] (xT) at (6,-2.4) {\footnotesize $X_T$};
    \path (h1) edge (o1);
    \path (h1) edge (h2);
    \path (o1) edge (h2);
    \path (h2) edge (o2);
    \path (h2) edge (hdots);
    \path (o2) edge (hdots);
    \path (hT) edge (oT);
    \path (hdots) edge (hT);
    \path (x1) edge (o1);
    \path (x2) edge (o2);
    \path (xT) edge (oT);
    \path (u2) edge (h2);
    \path (uT) edge (hT);
    \path (v1) edge (o1);
    \path (v2) edge (o2);
    \path (vT) edge (oT);
\end{tikzpicture}
\caption{The SCM with the comonotonic copula. The key change is that the noise nodes are now scalars as opposed to vectors, i.e., $(U_t, V_t)$ as opposed to $(\fU_t, \fV_t)$.}
\label{fig:SCMComono}
\end{figure}

As in \S\ref{sec:CopulaSimInd},  we start with the posterior samples $[h_{1:T}(b)]_{b=1}^B$ corresponding to the random path $H_{1:T} \mid (o_{1:T},  x_{1:T})$.
For each sample $b$,  our goal is to convert the sampled path $h_{1:T}(b)$ into a counterfactual path $\tildeh_{1:T}(b)$.
As noted in \S\ref{sec:CFModel}, irrespective of the copula choice, the counterfactual hidden state in period 1 equals the posterior sample, i.e.,
\begin{align*}
\tildeh_1(b) = h_1(b).
\end{align*}
To generate $\tildeo_1(b)$, we revisit the SCM in Figure \ref{fig:SCMComono}, which now has the noise nodes as scalars (as opposed to vectors). This is a direct implication of the comonotonic copula - see the statement immediately below (\ref{eq:CoM1}).

By the structural equation \eqref{eq:SCMone}, $\tildeo_1(b)$ equals
\begin{subequations}
\begin{align}
\tildeo_1(b) = f(\tildeh_1(b),  \tx_1,  V_1) = f_{\tildeh_1(b) \tx_1}(V_1),
\label{eq:CFOOne}
\end{align}
where $f_{hx}(\cdot)$ is the inverse transform function corresponding to the rank-ordered CDF $[E_{hx i}]_i$ (recall \eqref{eq:RankCDFE}).  Hence, all we need to sample $\tildeo_1(b)$ is the posterior distribution of $V_1$, where the ``posterior'' corresponds to conditioning on $O_{1 h_1(b) x_1} = o_1$ (recall the notation $O_{thx}$ from \S\ref{sec:opt}). Given the prior $V_1 \sim \text{Unif}[0,1]$,  we can compute the posterior in closed-form. In particular,
\begin{align}
V_1 \mid (O_{1 h_1(b) x_1} = o_1) \sim \text{Unif}[E_{h_1(b) x_1 o_1^{-}}, E_{h_1(b) x_1 o_1}],
\label{eq:VOnePost}
\end{align}
\end{subequations}
where $o^- := r_O^{-1}(r_O(o) - 1)$ is the emission ranked just below $o$. Hence, we can efficiently sample $V_1$ from its posterior,  and this $V_1$ sample can be used to generate $\tildeo_1(b)$ (via \eqref{eq:CFOOne}). Given we encoded rank orderings in the CDF $E_{hxi}$, such sampling will naturally enforce pathwise monotonicity.

We can sample $\tildeh_2(b)$ similarly.  By the structural equation \eqref{eq:SCMtwo}, $\tildeh_2(b)$ equals
\begin{subequations}
\begin{align}
\tildeh_2(b) = g(\tildeh_1(b),  \tildeo_1,  U_2) = g_{\tildeh_1(b) \tildeo_1}(U_2),
\label{eq:CFHTwo}
\end{align}
where $g_{hi}(\cdot)$ is the inverse transform function corresponding to the rank-ordered CDF $[Q_{hi h'}]_{h'}$ (recall \eqref{eq:RankCDFQ}).
Hence, all we need to sample $\tildeh_2(b)$ is the posterior distribution of $U_2$, where the ``posterior'' corresponds to conditioning on $H_{2 h_1(b) o_1} = h_2(b)$ (recall the notation $H_{thi}$ from \S\ref{sec:opt}).
Given the prior $U_2 \sim \text{Unif}[0,1]$,  we can compute the posterior in closed-form. In particular,
\begin{align}
U_2 \mid (H_{2 h_1(b) o_1} = h_2(b)) \sim \text{Unif}[Q_{h_1(b) o_1 h_2(b)^{-}}, Q_{h_1(b) o_1 h_2(b)}],
\label{eq:UTwoPost}
\end{align}
\end{subequations}
where $h^- := r_H^{-1}(r_H(h) - 1)$  is the state ranked just below $h$. Hence, we can efficiently sample $U_2$ from its posterior,  and this $U_2$ sample can be used to generate $\tildeh_2(b)$ (via \eqref{eq:CFHTwo}). Given we encoded rank orderings in the CDF $Q_{hih'}$, such sampling will naturally enforce pathwise monotonicity.

We then generate period 2 counterfactual emission $\tildeo_2(b)$ in a similar manner and the process repeats until we hit the end of horizon.  We summarize the procedure in Algorithm \ref{alg:CopulaSimComono}.

\begin{algorithm}[H]
\begin{algorithmic}[1]
\REQUIRE{$(\fE, \fQ)$,   $(o_{1:T}, x_{1:T})$,  $[h_{1:T}(b)]_{b=1}^B$, $\tx_{1:T}$, $r_H(\cdot)$, $r_O(\cdot)$}
\FOR{$b=1$ to $B$}
\STATE{$\tildeh_1(b) = h_1(b)$}
\FOR{$t=1$ to $T-1$}
\STATE{$v_t \sim \text{Unif}[E_{h_t(b) x_t o_t^{-}}, E_{h_t(b) x_t o_t}]$ \hfill \% posterior sample of $V_t$ (see \eqref{eq:VOnePost})}
\STATE{$\tildeo_t(b) = f_{\tildeh_t(b) \tx_t}(v_t)$ \hfill \% counterfactual emission  (see \eqref{eq:CFOOne})}
\STATE{$u_{t+1} \sim \text{Unif}[Q_{h_t(b) o_t h_{t+1}(b)^{-}}, Q_{h_t(b) o_t h_{t+1}(b)}]$ \hfill \% posterior sample of $U_{t+1}$ (see \eqref{eq:UTwoPost})}
\STATE{$\tildeh_{t+1}(b) = g_{\tildeh_t(b) \tildeo_t}(u_{t+1})$ \hfill \% counterfactual state (see \eqref{eq:CFHTwo})}
\ENDFOR
\ENDFOR
\STATE{\textbf{return} $[\tildeh_{1:T}(b)]_b$}
\end{algorithmic}
\caption{Counterfactual simulations under the comonotonic copula}
\label{alg:CopulaSimComono}
\end{algorithm}

\section{Enhancing the Scalability of the Polynomial Optimization}  \label{sec:scalability}

In this section, we discuss ways to enhance the scalability of the polynomial optimizations in \eqref{eq:MaxMin}. First, in \S\ref{sec:reformulationApp}, we show how the optimization can be reformulated to avoid the exponential dependence on $T$ (recall the discussion towards the end of \S\ref{sec:optProblem}).  Second, in \S\ref{sec:PairwiseOpt}, we discuss an approximate way to optimize our problem that drastically reduces the underlying dimensionality of the problem.
Third,  in \S\ref{sec:CompPerformance}, we combine our ideas from \S\ref{sec:reformulationApp} and \S\ref{sec:PairwiseOpt} and demonstrate (via numerics) that we can obtain high-quality solutions for $T$ as large as $100$ in just a few hours of compute time.

Related to scalability, we mention in passing that in each of our optimization problems,  we added the constraint that the objective value (which is a probability) must lie in $[0,1]$.  Of course, this constraint is redundant but we found it helped speed up the solver convergence in a few instances, possibly because it shrunk the search space as the solver does not know a priori that the objective is a probability.

\subsection{Reformulating the Polynomial Optimization to Avoid the Exponential Dependence on $T$}  \label{sec:reformulationApp}

Recall Lemmas \ref{lem:PNdecomp} and \ref{lem:PNrecursive}, which characterize the objective function of our polynomial optimization problem. We repeat them here for the sake of convenience.

\PNdecomp*

\PNrecursive*

It is easy to see that a naive expansion of PN (as per Lemmas \ref{lem:PNdecomp} and \ref{lem:PNrecursive}) results in a number of terms that is exponential in $T$.  This is clearly undesirable since we end up running into memory issues for even a moderate value of $T$.  For example,  such issues arise for $T > 10$ in the breast cancer numerics of \S\ref{sec:exp}.  It is possible to remove this exponential dependence, however, by a reformulation of the optimization, which we now discuss. (Note that the objective function remains the same irrespective of whether we optimize over the pairwise marginals (as discussed in \S\ref{sec:PairwiseOpt}) or the joint distribution (as presented in \S\ref{sec:optProblem}) and hence, the reformulation here is ``universal''.)

The reformulation steps are as follows:
\begin{enumerate}
\item Define $\Pb_{\tM(b)} (\tildeh_t)$ from Lemma \ref{lem:PNrecursive} as a decision variable for all $(t, \tildeh_t, b) \in [T] \times \bH \times [B]$.
\item Add the Lemma \ref{lem:PNrecursive} equations as constraints in the optimization (for each $(t, \tildeh_t, b) \in [T] \times \bH \times [B]$). Note that these are non-linear but polynomial constraints and hence, we remain within the class of polynomial programs.  Furthermore, none of the constraints have an exponential number of terms since $\Pb_{\tM(b)} (\tildeh_t)$ are decision variables now.
\item The objective now is simply the expression in Lemma \ref{lem:PNdecomp}.
\end{enumerate}

These steps result in the following\footnote{Note that we focus on the maximization problem from \eqref{eq:MaxMin} but the same holds for the minimization counterpart. All we need to do is simply change the ``max'' to a ``min'' in the objective function \eqref{eq:reformulationObj}.} optimization, where we use the decision variable $\gamma^t_{\tildeh_t b}$ to denote the probability term\footnote{To be pedantic, we could have added a ``$t$'' super-script in $\Pb_{\tM(b)} (\tildeh_t)$ and used the notation $\Pb^{t}_{\tM(b)} (\tildeh_t)$ instead. However,  we did not do so earlier since this dependence on $t$ was implicitly understood to exist, and adding this extra super-script felt unnecessary.} $\Pb_{\tM(b)} (\tildeh_t)$ in the LHS of Lemma \ref{lem:PNrecursive} for all $(t, \tildeh_t, b) \in [T] \times \bH \times [B]$,  with $\pmb{\gamma} := [\gamma_{hb}^t]_{(t,h,b)}$:
\begin{restatable}{subequations}{reformulation} \label{eq:reformulation}
\begin{align}
\max_{(\ftheta, \fpi) \in \cF, \pmb{\gamma} }\ & \left\{1 - \frac{1}{B} \sum_{b=1}^B \gamma^T_{7B} \right\} \label{eq:reformulationObj} \\
\text{s.t.} \ &  \gamma_{hb}^t = \sum_{h' \in \bH} \sum_{o' \in \bO} \frac{\pi_{h' o', h_{t-1}(b) o_{t-1}}(h, h_{t}(b))}{q_{h_{t-1}(b) o_{t-1} h_{t}(b)}} \times \frac{\theta_{h' \tx_{t-1}, h_{t-1}(b) x_{t-1}}(o',  o_{t-1})}{e_{h_{t-1}(b) x_{t-1} o_{t-1}}} \times \gamma_{h'b}^{t-1} &\forall t > 1 \ \forall h \in \bH \ \forall b \in [B]  \\
& \gamma^1_{hb} = 1 & \forall h = h_1(b) \ \forall b \in [B] \\
& \gamma^1_{hb} = 0 & \forall h \neq h_1(b)\ \forall b \in [B].
\end{align}
\end{restatable}
As before (refer to \S\ref{sec:opt}), the feasibility set $\cF$ over $(\ftheta,\fpi)$ can correspond to \eqref{eq:marginalconsthetapi}, \eqref{eq:pairwisemarginallink},  and \eqref{eq:nonnegativityjoint}.  It can also include additional constraints such as CS and PM, or correspond to the lower-dimensional space over the pairwise marginals (as discussed in \S\ref{sec:PairwiseOpt}).
Clearly,  \eqref{eq:reformulation} has a linear objective and polynomial constraints, and is therefore also a polynomial program. The number of terms in the objective is no longer exponential in $T$ but this has come at the cost of having to add a total of $\lvert \bH \rvert T B$ decision variables and (polynomial) constraints to the original formulation in \eqref{eq:MaxMin}.  Though the size of our reformulation (number of variables and constraints) scales with both $T$ and $B$, we found it to scale much more gracefully (with respect to $T$) than the original formulation, as we discuss in \S\ref{sec:CompPerformance} below.

Note that we do not necessarily need to add these $\lvert \bH \rvert T B$ variables and constraints to the optimization but for that, we need the ability to modify the source code of the optimization solver (\texttt{BARON} in our case). This is because even in the original formulation \eqref{eq:MaxMin}, we can actually evaluate the objective function in polynomial time and space rather than naively expanding it into exponentially many terms. To see this, consider a given sample number $b \in [B]$. We need to evaluate $\Pb_{\tM(b)} (\tH_T = 7)$ from Lemma \ref{lem:PNrecursive}. To do so, we start from period 1 and store $\Pb_{\tM(b)} ( \tildeh_1 )$ for all $\tildeh_1 \in \bH$ (see Lemma \ref{lem:PNrecursive}'s base case). We then move to period 2 and store $\Pb_{\tM(b)} ( \tildeh_2 )$ for all $\tildeh_2 \in \bH$ (see Lemma \ref{lem:PNrecursive}'s recursion).  The key here is that when computing $\Pb_{\tM(b)} ( \tildeh_2 )$, we make use of the stored values of $\Pb_{\tM(b)} ( \tildeh_1 )$.  We then move to period 3 evaluations, where we make use of the stored values of $\Pb_{\tM(b)} ( \tildeh_2 )$.  We repeat this procedure until we hit period $T$.  Clearly, this procedure requires polynomial time and space.  Furthermore, we can evaluate the gradient (and the Hessian) of $\Pb_{\tM(b)} (\tH_T = 7)$ in a similar manner (if needed by the optimization solver).  We can therefore evaluate the objective and its gradient information at a given point in polynomial time and space. These can then be used by the optimization solver. However, we are unable to modify the solver we use (\texttt{BARON}), and \texttt{BARON} by default does not exploit this structure but naively expands the objective into $\exp(T)$ terms.  As such, we use the reformulation presented in \eqref{eq:reformulation} instead.

\subsection{Approximating the Joint Optimization by the Pairwise Optimization}  \label{sec:PairwiseOpt}

The problem \eqref{eq:MaxMin} discussed in \S\ref{sec:optProblem} optimizes over the joint PMFs (``joint optimization''). The challenge here lies in the dimensionality of the underlying joint distribution.  As discussed towards the end of \S\ref{sec:optProblem}, the problem size (number of decision variables in particular) can grow exponentially in the primitives (e.g.,  $\lvert \bH \rvert$, $\lvert \bX \rvert$, and $\lvert \bO \rvert$).
This is because the decision variables capture the entire joint distribution.
Though we might be able to exploit application-specific sparsity to manage this blow-up (as we in fact do for the breast cancer application), it is worth exploring if there is a more tractable alternative in general (i.e., not specific to any application).  We now show that this is possible.

Recall from \S\ref{sec:optProblem} that we are interested in the following optimizations (repeating \eqref{eq:MaxMin} for convenience):
\MaxMin*
The key observation here is that the objective function does not depend on the joint PMF of $(\ftheta, \fpi)$ but only the corresponding pairwise marginals (recall Lemmas \ref{lem:PNdecomp} and \ref{lem:PNrecursive}).  We introduced the joint PMF decision variables to ensure the feasibility set $\cF$ is such that the pairwise marginals are valid.  However, as an alternative,  we can choose to not introduce the joint variables in the optimization and instead approximate $\cF$ by expressing it in terms of the pairwise variables. For example,  since the pairwise variables correspond to the 2-dimensional PMFs, they must obey basic probability axioms.  In particular, they must be non-negative and agree with their known 1-dimensional marginals so that
\begin{subequations}
\label{eq:marginalconstraints}
\begin{align}
\sum_{h'} \pi_{\tildeh \ti, hi}(\tildeh', h') &= q_{\tildeh \ti \tildeh'} \ \forall (\tildeh, \ti,  \tildeh') \ \forall (h,i) \\
\sum_{\tildeh'} \pi_{\tildeh \ti, hi}(\tildeh', h') &= q_{h i h'} \ \forall (\tildeh, \ti) \ \forall (h,i, h') \\
\sum_{i} \theta_{\tildeh \tx,  hx}(\ti, i) &= e_{\tildeh \tx \ti} \ \forall (\tildeh, \tx, \ti) \ \forall (h,x) \\
\sum_{\ti} \theta_{\tildeh \tx,  hx}(\ti, i) &= e_{h x i} \ \forall (\tildeh, \tx) \ \forall (h,x, i).
\end{align}
\end{subequations}
These constraints are analogous to \eqref{eq:marginalconsthetapi} in \S\ref{sec:optProblem}.  It is easy to see that if \eqref{eq:marginalconsthetapi} is obeyed, then so is \eqref{eq:marginalconstraints}. However, the reverse implication does not hold,  meaning the feasibility space defined by \eqref{eq:marginalconstraints} and non-negativity (say $\cF'$) is a super-set of the feasibility space $\cF$ in \S\ref{sec:optProblem}.
In other words,  though the constraints in $\cF'$ are necessary, they are not sufficient to ensure the pairwise marginals correspond to a valid joint distribution. Hence, optimizing over $\cF'$ (``pairwise optimization'')\footnote{Note that the pairwise optimization is identical to the joint optimization \eqref{eq:MaxMin} but with the following two differences: (a) $\cF$ replaced by $\cF'$ and (b) joint decision variables not defined.} is a relaxation to the problem of optimizing over $\cF$.  In fact, as we show via a simple example next, this relaxation can be strict.  (We thank an anonymous reviewer for this example.)

\begin{ex} \label{ex:PairwiseNotSufficient}
Consider three random variables $X$, $Y$, and $Z$ with the following pairwise marginals:
\begin{align*}
(X,Y) = \begin{cases}
(0,0) \text{ w.p. } 1/2 \\
(1,1) \text{ w.p. } 1/2
\end{cases}
\quad
(Y,Z) = \begin{cases}
(0,0) \text{ w.p. }1/2 \\
(1,1) \text{ w.p. } 1/2
\end{cases}
\quad
(X,Z) = \begin{cases}
(1,0) \text{ w.p. } 1/2 \\
(0,1) \text{ w.p. } 1/2. \\
\end{cases}
\end{align*}
It is easy to verify these pairwise marginals obey \eqref{eq:marginalconstraints} along with non-negativity. However, they do not correspond to any valid joint distribution over $(X,Y,Z)$.  To see this, suppose $(X,Y)$ realizes a value of $(0,0)$. Then, the pairwise marginal of $(Y,Z)$ implies $(Y,Z)$ has to be $(0,0)$, which implies $(X,Z)$ must be $(1,0)$, resulting in a contradiction. Therefore, the bivariate marginals are not consistent with any valid 3-dimensional joint distribution.
\end{ex}

Despite the relaxation being strict\footnote{Since the pairwise optimization is a relaxation of the joint optimization, it follows that Proposition \ref{prop:PNBounds} still holds for the bounds produced by the pairwise optimization.}, we found it to produce high-quality solutions and be highly scalable (discussed in \S\ref{sec:CompPerformance}). The high scalability is primarily driven by the lower dimensionality of the decision variables. In particular, the pairwise optimization has \emph{at most} $\lvert \bH \rvert^4 \lvert \bO \rvert^2 + \lvert \bH \rvert^2 \lvert \bO \rvert^2 \lvert \bX \rvert^2$ decision variables (recall that the joint optimization has an additional $\lvert \bO \rvert^{\lvert \bH \rvert \lvert \bX \rvert} + \lvert \bH \rvert^{\lvert \bH \rvert \lvert \bO \rvert}$ decision variables).  In fact,  after exploiting the sparsity in the breast cancer application (along with the variable elimination discussed in Footnote \ref{ft:elimination}), the pairwise optimization has only 1082 decision variables. This is in contrast to the joint optimization which has 16,124 decision variables.  Though the pairwise optimization has more constraints than the joint optimization, the difference is not that stark (2085 vs.\ 610). (The numbers reported here correspond to the objective formulation presented in \S\ref{sec:optProblem} as opposed to the reformulation in \S\ref{sec:reformulationApp}. The reformulation adds a total of $\lvert \bH \rvert T B$ decision variables and $\lvert \bH \rvert T B$ constraints to both the pairwise and the joint optimizations.)

\subsection{Computational Performance}  \label{sec:CompPerformance}

\begin{figure}[H]
  \begin{center}
    \subfigure[Compute time]{\label{fig:TimeLargeTPathOne}\includegraphics[width=.4\linewidth]{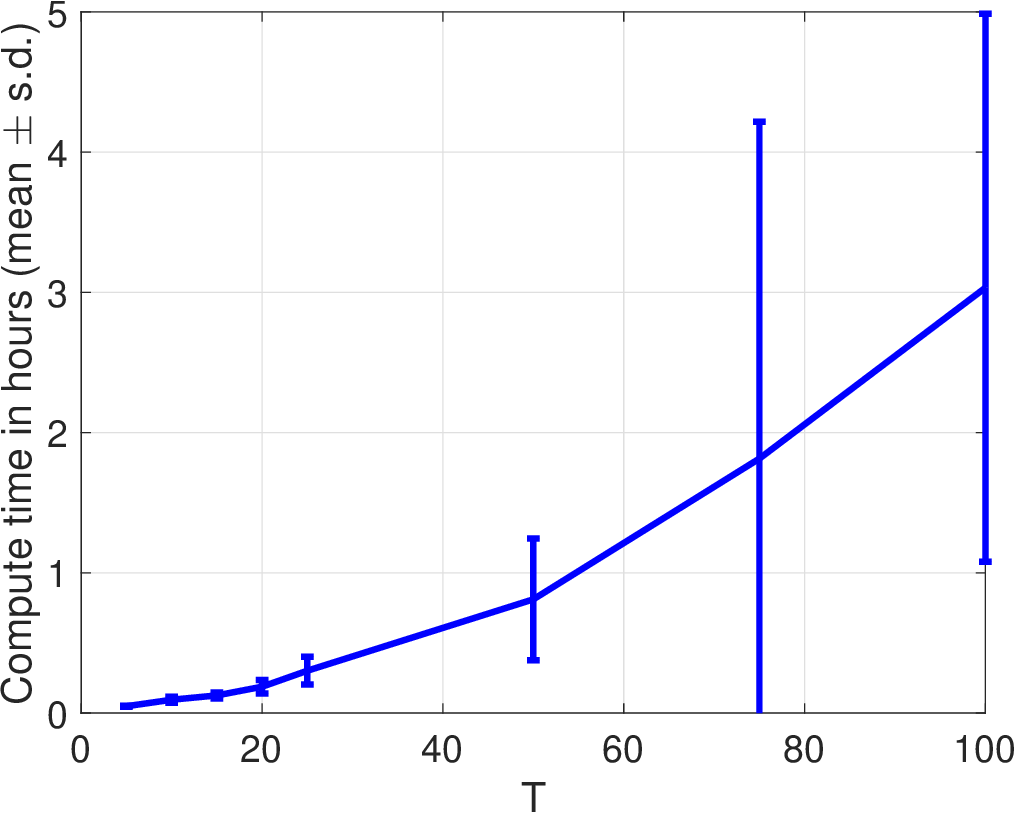}}
    \qquad
    \subfigure[PN]{\label{fig:PNLargeTPathOne}\includegraphics[width=.4\linewidth]{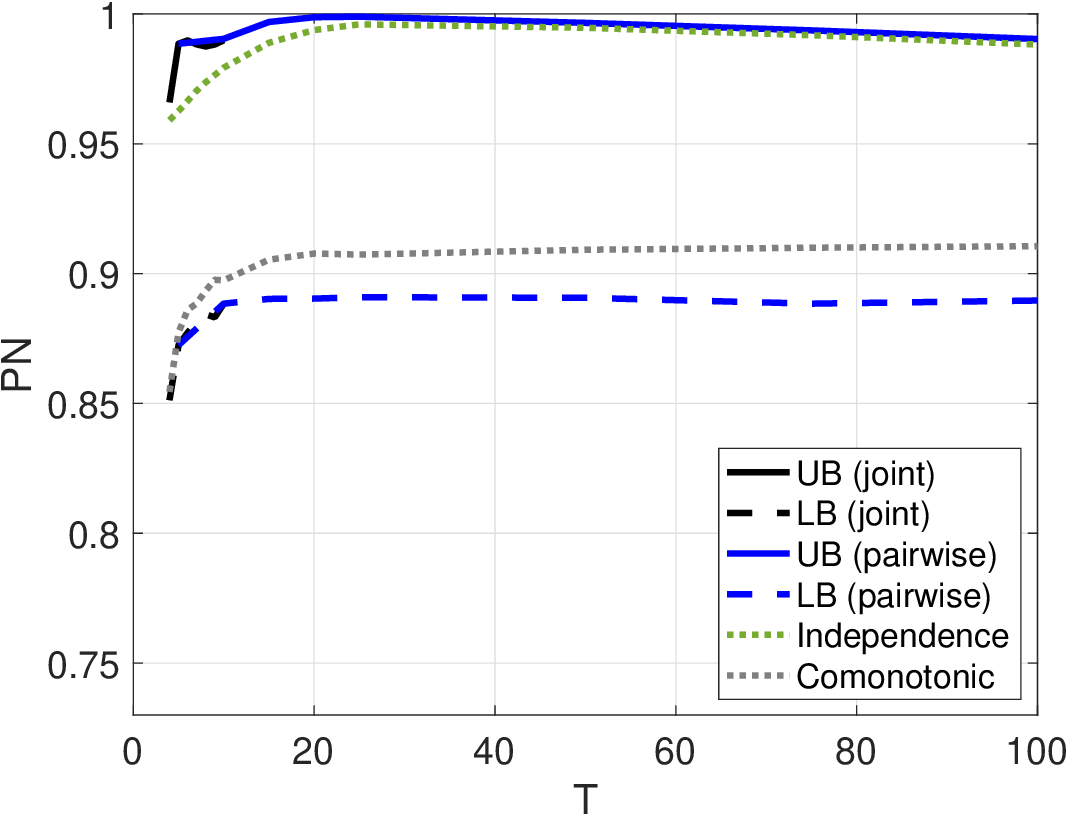}}
  \end{center}
\caption{Evaluating the computational performance of our ideas in \S\ref{sec:reformulationApp} and \S\ref{sec:PairwiseOpt} on path 1 from \S\ref{sec:exp}.  All results are averaged over the 20 seeds we use.  In sub-figure (b), the UB and LB curves for the ``joint'' optimization (black color) are the same as the ones in Figure \ref{fig:PN3a}, and only go as far as $T=10$ (since we run into memory issues for $T>10$). Furthermore,  to avoid clutter, we do not show the standard deviation bars in sub-figure (b) and note that the maximum standard deviation value is less than 0.01. To be clear, the compute times in sub-figure (a) correspond to the blue curves in sub-figure (b) (``pairwise''), which are the focus of this section.}
\label{fig:EfficientOptPathOne}
\end{figure}

We now compute upper and lower bounds on PN by (a) using the reformulation discussed in \S\ref{sec:reformulationApp} and (b) optimizing over the relaxed constraint set defined by the pairwise marginals as discussed in \S\ref{sec:PairwiseOpt}.\footnote{We also experimented with other variations of these two approaches.  If we use neither of them (as is the case in \S\ref{sec:exp}), then we run into memory issues for $T>10$.  In fact, even if we only use the second approach (optimizing over the relaxed constraint set), then we run into memory issues for $T>10$ since the objective still scales exponentially in $T$. Finally, if we only use the first approach, i.e. the reformulation of \S\ref{sec:reformulationApp}, and optimize over the joint, the \texttt{BARON} solver does not converge even for $T$ as small as $5$ in $24$ hours of compute time. This is because keeping the joint variables while doing the reformulation results in an optimization with a very large number of variables and constraints (even after we exploit sparsity).}
We focus on path 1 from \S\ref{sec:exp} for brevity and note that the results for path 2 are similar.  The implementation details remain the same as in \S\ref{sec:exp} (i.e., we code in \texttt{MATLAB-BARON} with \texttt{CPLEX} as the LP / MIP solver,  set absolute termination tolerance at $0.01$,  generate $B=100$ samples for SAA, and average over 20 seeds).  To test the scalability of our approach, we now experiment with $T \in \{5, 10,15,20,25,50,75, 100\}$. Note that $T=100$ is an order of magnitude larger than the longest horizon we have in \S\ref{sec:exp}, i.e., $T=10$. We next discuss the results that are shown in Figure \ref{fig:EfficientOptPathOne}, with Figure \ref{fig:TimeLargeTPathOne} showcasing scalability and Figure \ref{fig:PNLargeTPathOne} quality.

In Figure \ref{fig:TimeLargeTPathOne}, we display the compute time as a function of $T$. Compute time refers to the total time taken to compute LB and UB. Note that we let the solver run until convergence to global optimality (on just one core with at most 16 GB RAM).
We are able to solve for $T=100$ in 3 hours on average (over 20 seeds), with the minimum time being 1.2 hours and the maximum time being 8.4 hours.  This demonstrates the scalability of our approach.
(It is worth mentioning that after eliminating redundant variables and constraints, exploiting sparsity, using the reformulation of \S\ref{sec:reformulationApp}, and the pairwise approximation of \S\ref{sec:PairwiseOpt}, the $T=100$ and $B=100$ optimization has $71,082$ decision variables, $2085$ linear constraints, and $70,000$ polynomial constraints.)

In Figure \ref{fig:PNLargeTPathOne},  we display the PN values as a function of $T$. The ``joint'' UB and LB curves are the same as the ones in Figure \ref{fig:PN3a}, and only go as far as $T=10$ because of the aforementioned memory issues for $T>10$. The ``pairwise'' UB and LB curves are the focus of this section
and our goal here is to evaluate the quality of the ``pairwise'' bounds (blue curves) and we do so in two ways.
First,  as we are able to solve the joint optimization for $T \le 10$, we can use the ``joint'' bounds as benchmarks for the ``pairwise'' bounds. As may be seen from the figure,  the joint and pairwise bounds are very close to each other (for values of $T \leq 10$).
In particular, the joint and pairwise lower bounds coincide and equal $0.8725$ and $0.8884$ for $T=5$ and  $T=10$, respectively. The pairwise and joint upper bounds also coincide and equal $0.9885$ for $T=5$. The only difference between the two is the upper bound for $T=10$ with values of 0.9904 and 0.9899, respectively.
We therefore conclude that  the pairwise bounds provide a very good approximation to the joint bounds, at least when $T \le 10$.

Second, for $T > 10$, we use the fact that we can simulate the independence and comonotonic copulas, which by definition are feasible solutions to the joint optimization.  We therefore know that maximizing (minimizing) over the joint distribution will yield an upper (lower) bound that is no lower (higher) than the independence (comonotonic) curves in the figure.  As an example,  the gap between the independence and the pairwise UB curves (for $T > 10$) is never greater than $0.01$,  which means we lose at most $0.01$ by restricting ourselves to the pairwise marginals. Similarly,  the maximum gap between the comonotonic and the pairwise LB curves is $\sim 0.02$. Thus,  the pairwise bounds provide a high quality approximation to the joint bounds even when $T > 10$.

\begin{figure}[H]
  \begin{center}
    \subfigure[Counterfactual stability]{\label{fig:plot_PN_largeT_CS3}\includegraphics[width=.4\linewidth]{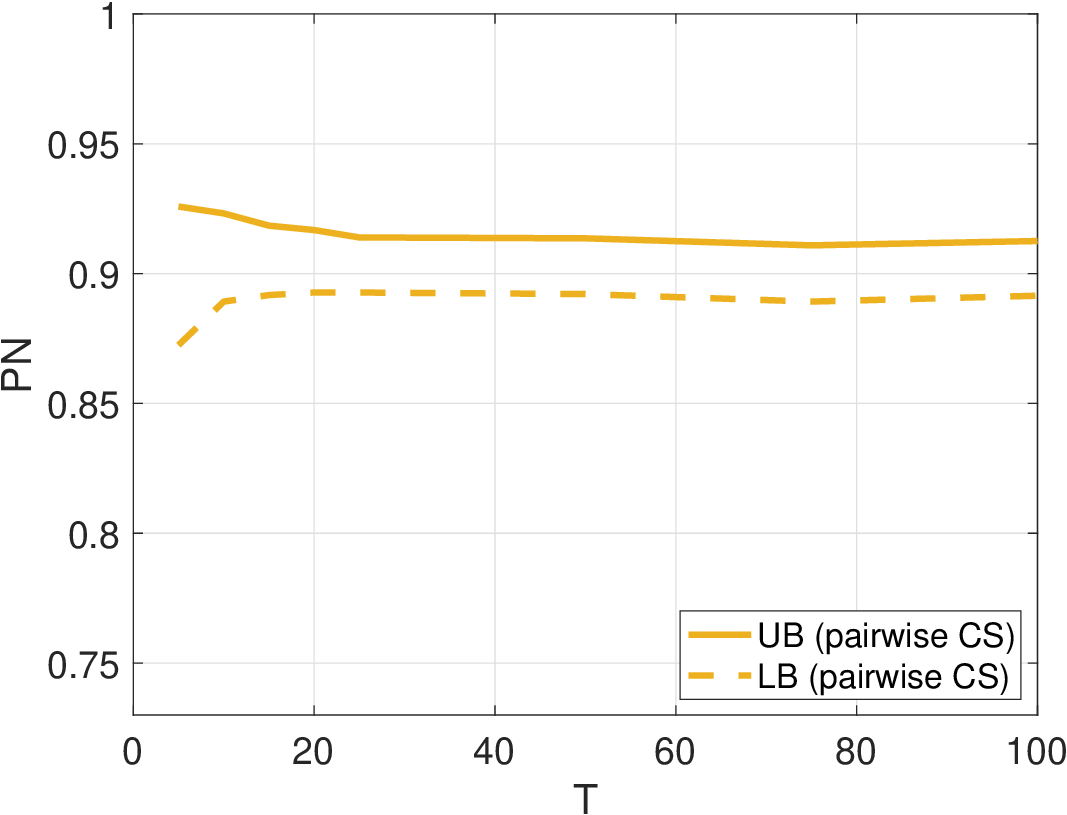}}
    \qquad
    \subfigure[Pathwise monotonicity]{\label{fig:plot_PN_largeT_PM3}\includegraphics[width=.4\linewidth]{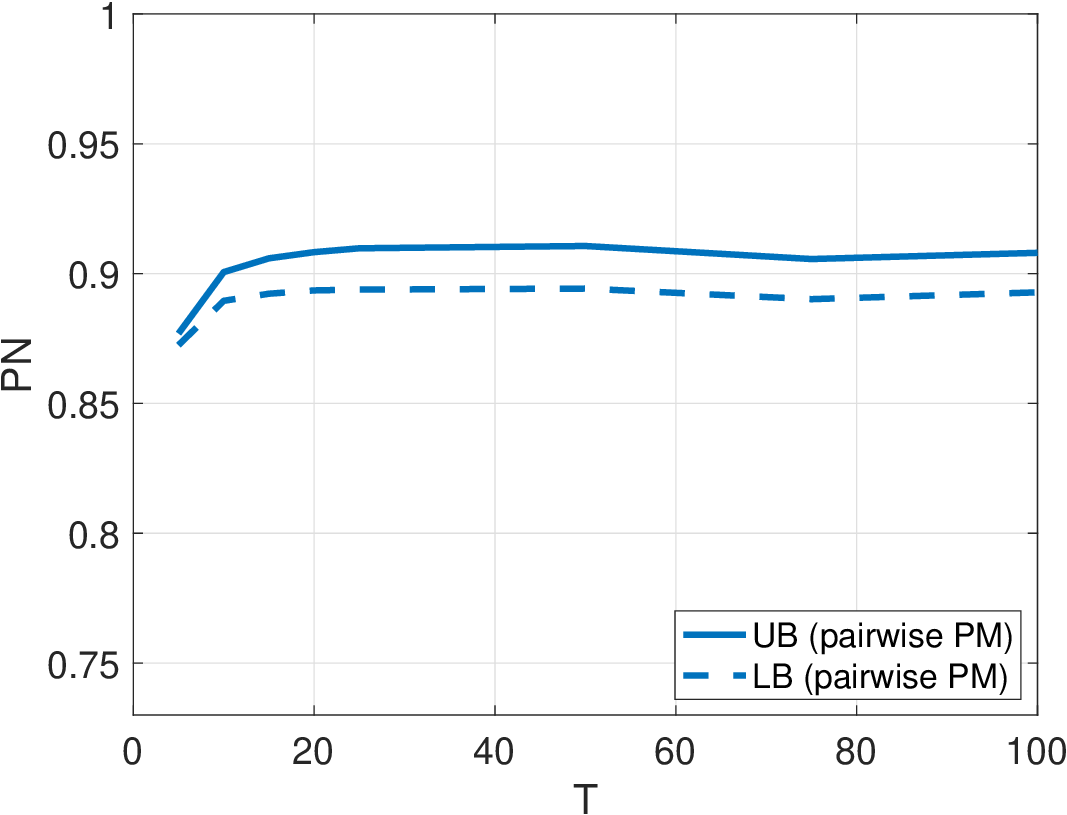}}
  \end{center}
\caption{PN bounds obtained when we embed CS and PM constraints in the pairwise optimization. All results are computed as the average of bounds obtained from 20 different seeds with each seed being used to generate $B=100$ paths.  Furthermore,  to avoid clutter, we do not show the standard deviation bars and note that the maximum standard deviation value is less than 0.01. }
\label{fig:EfficientOptPathOneCSPM}
\end{figure}

We can also embed CS and PM constraints in the pairwise optimization (recall from \S\ref{sec:optProblem} that both these constraints are over the pairwise variables) and we show the corresponding bounds in Figure \ref{fig:EfficientOptPathOneCSPM}. Naturally, the  bounds we obtain are tighter than the pairwise bounds in Figure \ref{fig:PNLargeTPathOne}. In particular, the UB gets much tighter while the LB does not change much.

\section{Further Details on the Breast Cancer Case Study}  \label{sec:expApp}
We discuss the breast cancer model primitives and their calibration in \S\ref{sec:CancerHMMPrimitives}, followed by showing how we exploit sparsity to reduce the number of decision variables (\S\ref{sec:jointsparsity}).
We then provide details on the PM constraints and the comonotonic copula in \S\ref{sec:PMCancer} and \S\ref{sec:comonocancer}, respectively. Finally,  in \S\ref{sec:pathtworesults},  we show the results for path 2.

\subsection{Model Primitives and Their Calibration} \label{sec:CancerHMMPrimitives}

As discussed in \S\ref{sec:problem}, the breast cancer application has $\lvert \bH \rvert = 7$ states,  $\lvert \bO \rvert = 7$ emissions, and $\lvert \bX \rvert = 2$ actions.
To be consistent with the literature \citep{ayer2012or},  we treat each period as corresponding to 6 months.
The model comprises of three primitives: $\fp$, $\fQ$, and $\fE$. We discuss their (sparse) structure and the calibration to real-data in \S\ref{sec:CancerPrimitiveP},  \S\ref{sec:CancerPrimitiveQ}, and \S\ref{sec:CancerPrimitiveE}, respectively.

\subsubsection{Initial State Distribution $\fp$} \label{sec:CancerPrimitiveP}

We have $\fp := (p_{1},  \ldots, p_{7})$ where $p_{h} := \Pb(H_1 = h)$ for all $h$. Usually, breast cancer screening starts around the age of 40 and the prevalence among females aged 40-49 is $1.0183\%$ (Table 4.24 of \citet{2017SEER},  all races, females): $$p_2 + p_3 = 0.010183.$$
Since in-situ cancer comprises $20\%$ of new breast cancer diagnoses \citep{sprague2009prevalence}, we get
\begin{align*}
p_2 &= 0.2 \times 0.010183 \\
p_3 &= 0.8 \times 0.010183.
\end{align*}
It is natural to set $$p_4 = p_5 = p_6 = p_7 = 0$$ and hence,  $$p_1 = 1 - p_2 - p_3 = 1 - 0.010183.$$

\subsubsection{Transition Distribution $\fQ$} \label{sec:CancerPrimitiveQ}

We have $\fQ := [q_{hih'}]_{h,i,h'}$ with $q_{hih'} := \Pb(H_{t+1} = h' \mid H_t = h, O_t=i)$.  Before discussing the calibration, we discuss the sparse structure of $\fQ$. To do so, we define the transition matrix $\fQ(i) := [q_{hih'}]_{hh'}$ for each emission $i$ (so that each row sums to 1) and observe that we have the following structure:

\[
\fQ(1) = \left[
\begin{array}{ccccccc}
\q_{11} & \q_{12}          & \q_{13}          &        &        &         & \\
       & \q_{22}  & \q_{23}        &        &        &         & \q_{27} \\
       &                 & \q_{33}  &        &        &         & \q_{37} \\
       &                 &                 & \q_{44} & \q_{45} & \q_{46}  & \q_{47}  \\
       &                 &                 &        & \q_{55} & \q_{56}  & \q_{57}\\
       &  &  &   &    &        1 & \\
       &  &  &   &    &         & 1 \\
\end{array}
\right]
\hspace{.5cm}
\fQ(2) = \left[
\begin{array}{ccccccc}
\q_{11} & \q_{12}          & \q_{13}          &        &        &         & \\
       & \q_{22}  & \q_{23}          &        &        &         & \q_{27} \\
       &                 & \q_{33}  &        &        &         &  \q_{37} \\
       &  &  &   &    &         & \\
       &  &  &   &    &         & \\
       &  &  &   &    &         & \\
       &  &  &   &    &         & \\
\end{array}
\right]
\]

\[
\fQ(3) = \left[
\begin{array}{ccccccc}
\q_{11} & \q_{12}          & \q_{13}          &        &        &         & \\
       &  &  &   &    &         & \\
       &  &  &   &    &         & \\
       &  &  &   &    &         & \\
       &  &  &   &    &         & \\
       &  &  &   &    &         & \\
       &  &  &   &    &         & \\
\end{array}
\right]
\hspace{0.5cm}
\fQ(4) = \left[
\begin{array}{ccccccc}
       &  &  &   &    &         & \\
       &  &  &  \q_{24}  &   \q_{25} & \q_{26} & \bar{\q}_{27}  \\
       &  &  &   &    &         & \\
       &                 &                 & \q_{44} & \q_{45} & \q_{46} & \q_{47} \\
       &  &  &   &    &         & \\
       &  &  &   &    &         & \\
       &  &  &   &    &         & \\
\end{array}
\right]
\]

\[
\fQ(5) = \left[
\begin{array}{ccccccc}
       &  &  &   &    &         & \\
       &  &  &   &    &         & \\
       &    &  &    & \q_{35} &  \q_{36}  & \bar{\q}_{37}  \\
       &  &  &   &    &         & \\
       &                 &                 &        & \q_{55} & \q_{56}  & \q_{57}\\
       &  &  &   &    &         & \\
       &  &  &   &    &         & \\
\end{array}
\right]
\hspace{.5cm}
\fQ(6) = \left[
\begin{array}{ccccccc}
       &  &  &   &    &         & \\
       &  &  &   &    &         & \\
       &  &  &   &    &         & \\
       &  &  &   &    &         & \\
       &  &  &   &    &         & \\
       &                 &                 &        &        & 1       & \\
       &  &  &   &    &         & \\
\end{array}
\right]
\hspace{.5cm}
\fQ(7) = \left[
\begin{array}{ccccccc}
       &  &  &   &    &         & \\
       &  &  &   &    &         & \\
       &  &  &   &    &         & \\
       &  &  &   &    &         & \\
       &  &  &   &    &         & \\
       &  &  &   &    &         & \\
       &                 &                 &        &        &         & 1\\
\end{array}
\right].
\]

A few comments are in order. First, an empty row means it is an impossible $(h,i)$ combination. For example,  if the we observe an emission $i=3$ (i.e., a negative biopsy),  then the underlying patient state has to be healthy, i.e., $h \notin \{2,3,4,5,6,7\}$.  Thus, rows 2 to 7 are empty in $\fQ(3)$.

Second,  observe that there is a decent amount of overlap across $[\fQ(i)]_i$ in terms of the underlying parameters. For example, $\q_{11}$ corresponds to the probability a healthy patient stays healthy, which is independent of the emission being 1 (no test), 2 (negative test), or 3 (positive test but negative biopsy).  Hence, $\q_{11}$ appears in all three matrices $\fQ(1)$, $\fQ(2)$, and $\fQ(3)$.  Of course, if the emission is 4, 5, 6, or 7, then the patient can not be healthy and hence, the corresponding entry in matrices $\fQ(4)$, $\fQ(5)$,  $\fQ(6)$, and $\fQ(7)$ is absent (in fact, the entire first row is empty, which means it is an impossible $(h,i)$ combination as discussed above).

Third, some rows have only a partial set of entries, which means that the other entries equal 0. For example,  if a patient is healthy (state 1), then her state can not transition to 4 (diagnosed in-situ with treatment started), 5 (diagnosed invasive with treatment started), 6 (recovered), or 7 (death) and hence, $\q_{14} = \q_{15} = \q_{16} = \q_{17} = 0$.  Hence, we do not show $\q_{14}, \q_{15}, \q_{16}, \q_{17}$ in $\fQ(1)$, $\fQ(2)$, or $\fQ(3)$.

Fourth, observe that we have a ``bar'' over $\bar{\q}_{27}$ (in $\fQ(4)$) and $\bar{\q}_{37}$ (in $\fQ(5)$). This is done to recognize them being different from $\q_{27}$ (in $\fQ(1)$ and $\fQ(2)$) and $\q_{37}$ (in $\fQ(1)$ and $\fQ(2)$). To see the difference, consider $\q_{37}$ versus $\bar{\q}_{37}$.  $\q_{37}$ corresponds to the patient state transitioning from invasive cancer to death when the cancer was not detected (and hence,  no treatment). On the other hand, $\bar{\q}_{37}$ corresponds to the patient state transitioning from invasive cancer to death when the cancer was detected (and hence,  treatment was provided). Naturally, we expect $\bar{\q}_{37} \le \q_{37}$.

Finally, since states 6 (recovery) and 7 (death) are absorbing, we have $\q_{66} = \q_{77} = 1$.

Having discussed the structure of $\fQ$, we now calibrate it to real-data. We iterate over each state $h \in \{1, \ldots, 7\}$ in a sequential manner.

\paragraph{State 1 (healthy).} For state 1, we are interested in $(\q_{11}, \q_{12},  \q_{13})$. These probabilities can depend on a woman's age but we ignore that and work with averages.  Let's focus on $(\q_{12}, \q_{13})$ since $$\q_{11} = 1 - \q_{12} - \q_{13}.$$
For $\q_{12}$,  we use the in-situ incidence rates from Table 4.12 of \citet{2017SEER} (all races,  females).
For $\q_{13}$, we use the invasive incidence rates from Table 4.11 of \citet{2017SEER} (all races, females).
The reported numbers are per year and we should divide by 2 to convert to a 6-month scale:
\begin{align*}
\q_{12} &= \frac{1}{2} \times \frac{33.0}{100000} \\
\q_{13} &= \frac{1}{2} \times \frac{128.5}{100000}.
\end{align*}
Note that consistent with the 20-80 split in $(p_2, p_3)$,  we have $\q_{13} \approx 4 \q_{12}$.

\paragraph{State 2 (undiagnoised in-situ cancer).} We are interested in $(\q_{22},  \q_{23},  \q_{27})$ (if cancer is not detected) and $(\q_{24},  \q_{25}, \q_{26}, \bar{\q}_{27})$ (if cancer is detected). First,  consider $(\q_{22},  \q_{23}, \q_{27})$. Table 4.13 of \citet{2017SEER} and Page 26 of \citet{2013UWBCS} imply there is no death from in-situ cancer: $$\q_{27} = 0.$$
\citet{haugh2019information} assumed $\q_{23}$ to equal the invasive incidence rate $\q_{13}$ and so do we:
\begin{align*}
\q_{23} &= \q_{13} \\
\q_{22} &= 1 - \q_{23} - \q_{27} = 1 - \q_{13}.
\end{align*}
Second,  consider $(\q_{24},  \q_{25},  \q_{26}, \bar{\q}_{27})$. As $\q_{27} = 0$ and we expect $\bar{\q}_{27}  \le \q_{27}$ (recall comment \#4 above), we set $$\bar{\q}_{27} = 0.$$
As all in-situ cancer patients survive (if treated), no one transitions to invasive (if in-situ detected): $$\q_{25} = 0.$$
Finally, we have $$\q_{24} + \q_{26} = 1.$$
The split between $\q_{24}$ and $\q_{26}$ is irrelevant in terms of the patient dying or not (all will survive as there is no positive probability path from state 4 to death; this will become clear when we discuss state 4 below).

\begin{figure}[ht]
\centering
\includegraphics[width=0.4\linewidth]{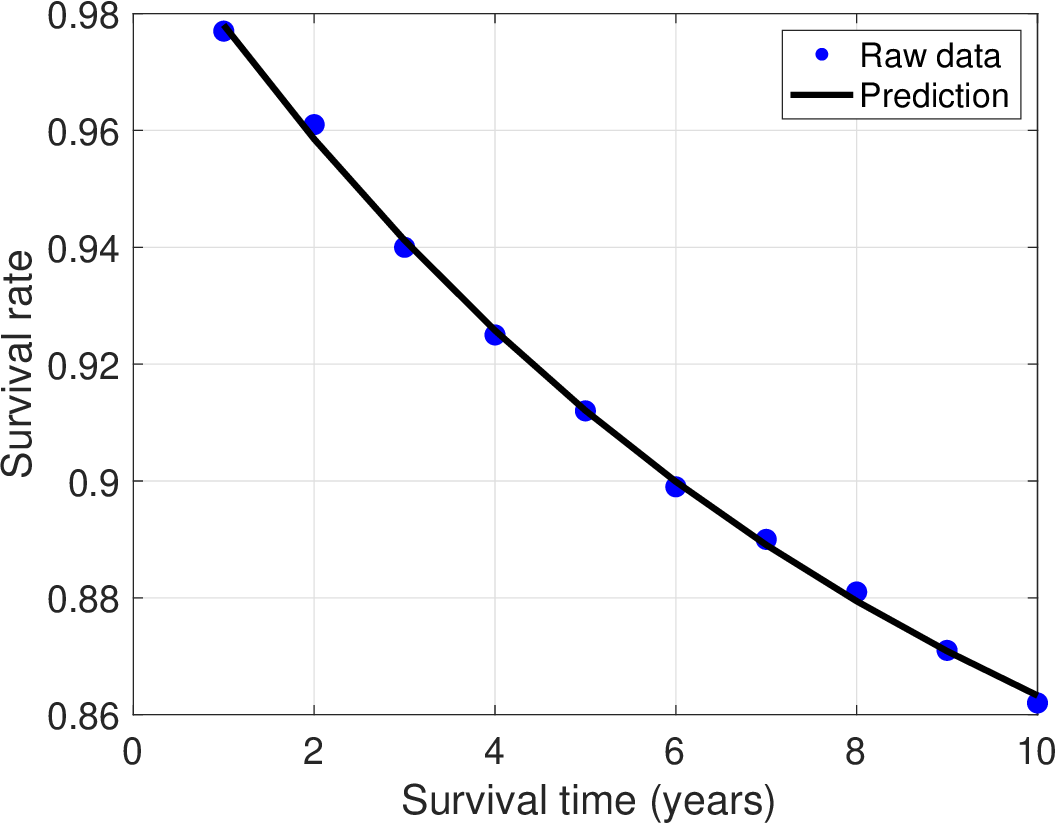}
\caption{Calibration of $(\q_{36}, \bar{\q}_{37})$. Under our (time-invariant) Markov model,  with a starting state of invasive breast cancer (under treatment), the survival rate after $x$ years equals $1 - \bar{\q}_{37} \sum_{i=0}^{2x-1} (1-\q_{36}-\bar{\q}_{37})^i $.  Minimizing the sum of squared errors (on the 10 blue data points in the plot) over $(\q_{36}, \bar{\q}_{37})$ gives us an estimate of $(0.0459,  0.0113)$. The prediction using our fit is shown via the black curve.}
\label{fig:invasivesurvival}
\end{figure}

\paragraph{State 3 (undiagnoised invasive cancer).} We are interested in $(\q_{33}, \q_{37})$ (if cancer is not detected) and $(\q_{35},  \q_{36}, \bar{\q}_{37})$ (if cancer is detected).  First, consider $(\q_{33}, \q_{37})$. $\q_{37}$ is the probability of dying from invasive breast cancer (under no treatment). According to \citet{johnstone2000survival}, the 5-year and 10-year survival rates for invasive breast cancer patients (under no treatment) are $18.4\%$ and $3.6\%$, respectively.  On calibrating to 5-year rate, we get $(1 - \q_{37})^{10} = 0.184$, which implies $\q_{37} \approx 15.6\%$ (note that we use ``10'' in the exponent since our time periods correspond to 6 months and and hence, 5 years correspond to 10 periods). Similarly,  on calibrating to 10-year rate, we get $(1 - \q_{37})^{20} = 0.036$ implies $\q_{37} \approx 15.3\%$. The two calibrations are consistent with each other (lending evidence to time-invariance). Minimizing sum of squared errors over the two data points, i.e., $\min_{\q_{37} \in [0,1]} \{((1 - \q_{37})^{10} - 0.184)^2 + ((1 - \q_{37})^{20} - 0.036)^2\}$, gives the following estimate:
\begin{align*}
\q_{37} &= 0.1554.
\end{align*}
Naturally, we have
\begin{align*}
\q_{33} &= 1 - \q_{37}.
\end{align*}
Second, consider $(\q_{35},  \q_{36}, \bar{\q}_{37})$. $\q_{36}$ and $\bar{\q}_{37}$ are the probabilities of recovering and dying from invasive breast cancer (under treatment). Table 4.14 of \citet{2017SEER} has various survival rates we can use to calibrate. We calibrate using the 10 data points corresponding to the year 2007 (see Figure \ref{fig:invasivesurvival}):
\begin{align*}
\q_{36} &= 0.0459 \\
\bar{\q}_{37} &= 0.0113.
\end{align*}
As a sanity check, note that $\bar{\q}_{37} < \q_{37}$. Finally, $$\q_{35} = 1 - \q_{36} - \bar{\q}_{37}.$$

\paragraph{State 4 (diagnoised in-situ cancer).} We are interested in $(\q_{44},  \q_{45},  \q_{46},  \q_{47})$.  Under our Markov model (which by definition is ``memoryless''),  it seems reasonable to set $$(\q_{44},  \q_{45},  \q_{46},  \q_{47}) = (\q_{24}, \q_{25}, \q_{26}, \bar{\q}_{27}).$$

\paragraph{State 5 (diagnoised invasive cancer).} We are interested in $(\q_{55}, \q_{56},  \q_{57})$. Under our Markov model,  it seems reasonable to set $$(\q_{55},  \q_{56},  \q_{57}) = (\q_{35},  \q_{36}, \bar{\q}_{37}).$$

\paragraph{States 6 (recovery) and 7 (death).} These two states are absorbing and hence, $$\q_{66} = \q_{77} = 1.$$

\subsubsection{Emission Distribution $\fE$} \label{sec:CancerPrimitiveE}

We have $\fE := [e_{hxi}]_{h,x,i}$ with $e_{hxi} := \Pb(O_{t} = i \mid H_t = h, X_t=x)$.  Before discussing the calibration, we discuss the sparse structure of $\fE$. To do so, we define the matrix $\fE(x) := [e_{hxi}]_{hi}$ for each action $x$ (so that each row sums to 1) and observe that we have the following structure:

\[
\fE(0) = \left[
\begin{array}{ccccccc}
0 & \e_{12}   & 1 - \e_{12} &        &        &        &      \\
0 & 1-\e_{24} &           & \e_{24} &        &        &        \\
0 & 1-\e_{35} &           &       & \e_{35} &        &       \\
0 &          &            & 1      &       &        &    \\
0 &          &            &        & 1       &        &    \\
0 &          &            &        &        &  1      &        \\
0 &          &            &        &        &        & 1        \\
\end{array}
\right]
\hspace{0.5cm}
\fE(1) = \left[
\begin{array}{ccccccc}
1 & 0 & 0 &   &   &   &      \\
1 & 0 & 0 &   &   &   &        \\
1 & 0 & 0 &   &   &   &       \\
0 & 0 & 0 & 1 &   &   &    \\
0 & 0 & 0 &   & 1 &   &    \\
0 & 0 & 0 &   &   & 1 &        \\
0 & 0 & 0 &   &   &   & 1    \\
\end{array}
\right].
\]

A few comments are in order. First,  for $x=1$ (no mammogram screening), the emission matrix $\fE(1)$ is extremely sparse with entries in $\{0,1\}$. For instance, when hidden state equals 1 (healthy), 2 (undiagnosed in-situ), or 3 (undiagnosed invasive), we observe no signal (emission equals 1) w.p.\ 1.  When hidden state equals 4, 5, 6, or 7, we naturally observe the same emission w.p.\ 1.

Second,  for $x=0$ (screening), the emission matrix $\fE(0)$ is quite sparse as well.  If the patient is healthy (row 1), then the test result is negative (true negative) w.p.\ $\e_{12}$ and positive (false positive) w.p.\ $1 - \e_{12}$.
When the patient has undiagnosed in-situ cancer (row 1), it is detected (true positive) w.p.\ $\e_{24}$ and missed (false negative) w.p.\ $1 - \e_{24}$.  The parameter $\e_{35}$ has the same interpretation as $\e_{24}$ but for invasive cancer. As for $x=1$,  when hidden state equals 4, 5, 6, or 7, we observe the same emission w.p.\ 1.

Having discussed the structure of $\fE$, we now calibrate it to real-data.  There are 3 parameters: $\e_{12}$, $\e_{24}$, and $\e_{35}$. All of them can be age specific but we ignore that. $\e_{12}$ is the \emph{specificity} of the mammogram screening (i.e., probability of a true negative) and we calibrate it using Table 3 of \citet{ayer2012or}:
\begin{align*}
\e_{12} = 0.9.
\end{align*}
$\e_{24}$ is the in-situ \emph{sensitivity} (i.e., probability of a true positive) and we calibrate it using Table 3 of \citet{ayer2012or}:
\begin{align*}
\e_{24} = 0.8.
\end{align*}
Finally, $\e_{35}$ is the invasive sensitivity and following \citet{ayer2012or}, we set $$\e_{35} = \e_{24}.$$

\subsection{Reducing the Number of Joint Decision Variables by Exploiting Sparsity} \label{sec:jointsparsity}
Recall from \S\ref{sec:optProblem} the following setup, which we repeat for convenience.  Let $k \equiv (h,x)$ and $m \equiv (h,i)$ so that
\begin{align*}
O_{k} &\equiv O_{hx}, \ e_{k i} \equiv e_{hxi} \\
H_m &\equiv H_{hi}, \ q_{mh'} \equiv q_{hih'}.
\end{align*}
We have $k \in [K]$ and $m \in [M]$, where $K := \lvert\bH\rvert \lvert \bX\rvert$ and $M := \lvert\bH\rvert \lvert \bO\rvert$.  The $K$ and $M$ dimensional joint PMFs for all $i_1, \ldots, i_K \in \bO$ and $h_1, \ldots, h_M \in \bH$ are defined as
\JointVariables*
As discussed towards the end of \S\ref{sec:optProblem},  it follows from \eqref{eq:JointVariables} that we have \emph{at most} $\lvert \bO \rvert^{\lvert \bH \rvert \lvert \bX \rvert} + \lvert \bH \rvert^{\lvert \bH \rvert \lvert \bO \rvert}$ joint variables. We now show that these are merely upper bounds and we can exploit the sparsity inherent in the underlying application to drastically reduce these numbers.

Consider the $\theta_{1,\ldots,K}$ decision variables for now.  Since $\theta_{1,\ldots,K}$ represents the joint PMF of the random variables $[O_k]_{k}$ where $k \equiv (h,x)$,  we first understand which $(h,x)$ pairs are valid (as opposed to naively considering all $(h,x) \in \bH \times \bX$).  Recall that the state $h$ has the following encoding:
\begin{enumerate}
\item healthy
\item undiagnosed in-situ cancer
\item undiagnosed invasive cancer
\item diagnosed in-situ cancer
\item diagnosed invasive cancer
\item recovery
\item death.
\end{enumerate}
Furthermore,  $x$ equals 0 maps to mammogram being performed and 1 to it not being performed. It is easy to see that all 14 combinations of $(h,x) \in \bH \times \bX$ are valid so none of the corresponding decision variables can be set to zero (and therefore removed). Turning now to the observations, we recall that they are encoded as follows:
\begin{enumerate}
\item no screening took place
\item negative screening result (possibly a false negative)
\item positive mammogram result, but followed by a negative biopsy
\item diagnosed in-situ cancer
\item diagnosed invasive cancer
\item recovery
\item death.
\end{enumerate}
Given this, Table \ref{tab:thetasparisty} documents the range of all $7 \times 2= 14$ random variables $[O_{hx}]_{h,x}$.
\begin{table}[H]
\caption{Range of the 14 random variables $[O_{hx}]_{h,x}$ corresponding to $\theta_{1,\ldots,K}$.}
\label{tab:thetasparisty}
\vskip 0.05in
\begin{center}
\begin{small}
\begin{tabular}{cccc}
\toprule
State $h$ & Policy $x$ & Range of $O_{hx}$ & Range cardinality  \\
\midrule
1     & 0 & $\{2,3\}$ & 2 \\
1     & 1 & $\{1\}$ & 1 \\
2    & 0 & $\{2,4\}$ & 2 \\
2    & 1 & $\{1\}$ & 1 \\
3    & 0 & $\{2,5\}$ & 2 \\
3    & 1 & $\{1\}$ & 1 \\
4    & 0 & $\{4\}$ & 1 \\
4    & 1 & $\{4\}$ & 1 \\
5    & 0 & $\{5\}$ & 1 \\
5    & 1 & $\{5\}$ & 1 \\
6    & 0 & $\{6\}$ & 1 \\
6    & 1 & $\{6\}$ & 1 \\
7    & 0 & $\{7\}$ & 1 \\
7    & 1 & $\{7\}$ & 1 \\
\bottomrule
\end{tabular}
\end{small}
\end{center}
\vskip -0.1in
\end{table}
Multiplying all of the 14 cardinalities (last column in Table \ref{tab:thetasparisty}) implies that there are only eight $\theta_{1,\ldots,K}$ decision variables that need to be considered. This is in contrast to the upper bound of $\lvert \bO \rvert^{\lvert \bH \rvert \lvert \bX \rvert} = 7^{14}$.

The same logic applies to the $\pi_{1,\ldots,M}$ decision variables. In fact, for the $\pi_{1,\ldots,M}$ decision variables, even the first step proves useful since not all $(h,i)$ pairs are valid. For instance, if $h=1$, then $i \notin \{4,5,6,7\}$. In particular, the first step allows us to trim down the number of $[H_{hi}]_{h,i}$ random variables from $\lvert \bH \rvert \lvert \bO \rvert = 49$ to $13$. The second step trims down the range of each of the $13$ random variables. We document this in Table \ref{tab:pisparisty} and are able to reduce the number of $\pi_{1,\ldots,M}$ decision variables from $\lvert \bH \rvert^{\lvert \bH \rvert \lvert \bO \rvert} = 7^{49}$ to $15,552$ (which equals the product of the cardinalities presented in the last column).

\begin{table}[H]
\caption{Range of the 13 random variables $[H_{hi}]_{h,i}$ corresponding to $\pi_{1,\ldots,M}$. Only 13 $(h,i)$ pairs are shown as the other 36 are not valid.}
\label{tab:pisparisty}
\vskip 0.05in
\begin{center}
\begin{small}
\begin{tabular}{cccc}
\toprule
State $h$ & Observation $i$ & Range of $H_{hi}$ & Range cardinality  \\
\midrule
1     & 1 & $\{1,2,3\}$ & 3 \\
1     & 2 & $\{1,2,3\}$ & 3 \\
1    & 3 & $\{1,2,3\}$ & 3 \\
2    & 1 & $\{2,3\}$ & 2 \\
2    & 2 & $\{2,3\}$ & 2 \\
2    & 4& $\{4,6\}$ & 2 \\
3    & 1 & $\{3,7\}$ & 2 \\
3    & 2 & $\{3,7\}$ & 2 \\
3    & 5 & $\{5,6,7\}$ & 3 \\
4    & 4 & $\{4,6\}$ & 2 \\
5    & 5 & $\{5,6,7\}$ & 3 \\
6    & 6 & $\{6\}$ & 1 \\
7    & 7 & $\{7\}$ & 1 \\
\bottomrule
\end{tabular}
\end{small}
\end{center}
\vskip -0.1in
\end{table}

\subsection{Details on the Pathwise Monotonicity (PM) Constraints} \label{sec:PMCancer}
PM can be enforced via linear constraints.  We briefly discussed this in \S\ref{sec:optProblem} and now discuss all underlying PM constraints we embedded in our breast cancer numerics.

Recalling our \S\ref{sec:optProblem} discussion for convenience,  suppose the patient has in-situ cancer in period $t$ which is not detected but the patient's state remains at in-situ in period $t+1$. Then, in the counterfactual world, if the cancer is detected in period $t$, then PM would require that the cancer can not be worse than in-situ in period $t+1$, i.e.,
\begin{align*}
\bP(H_{\tildeh \ti} = \tildeh' \mid H_{hi} = h') = 0
\end{align*}
for $h=2$, $i \in \{1,2\}$, $h'=2$, $\tildeh \in \{2,4\}$, $\ti=4$, $\tildeh' \in \{5,7\}$.
There can be multiple such cases to consider and we can enforce all the PM constraints by setting the corresponding $\pi_{\tildeh \ti, h i}(\tildeh', h')$ variables equal to 0 as $\pi_{\tildeh \ti, h i}(\tildeh', h') = \bP(H_{hi} = h') \bP(H_{\tildeh \ti} = \tildeh' \mid H_{hi} = h')$.

Hence,  to provide details on which all PM constraints we enforce, it suffices to enumerate the $(h,i,h',\tildeh,\ti,\tildeh')$ combinations for which we set the $\pi_{\tildeh \ti, h i}(\tildeh', h')$ variables equal to 0.
To do so, we iterate over each state $h \in \{1,\ldots,7\}$.
(Note that for PM, there are no $(h,x,i,\tildeh,\tx,\ti)$ combinations for which we set the $\theta_{\tildeh \tx, h x}(\ti,  i)$ variables equal to 0.)

\paragraph{State $h=1$ (healthy).} We enforce PM for the following combinations:
\bi
\item If $(h,i,h')$ equals (healthy, whatever emission, healthy), then the counterfactual state $\tildeh'$ can not be in-situ, invasive, or death if $\tildeh$ is healthy. That is, $h = 1$, $i \in \bO$, $h' = 1$, $\tildeh=1$, $\ti \in \bO$, and $\tildeh' \in \{2,3,4,5,7\}$.
\item If $(h,i,h')$ equals (healthy, whatever emission, in-situ), then the counterfactual state $\tildeh'$ can not be healthy, invasive, or death if $\tildeh$ is healthy. That is, $h = 1$, $i \in \bO$, $h' = 2$, $\tildeh=1$, $\ti \in \bO$, and $\tildeh' \in \{1,3,5,7\}$.
\item If $(h,i,h')$ equals (healthy, whatever emission, invasive), then the counterfactual state $\tildeh'$ can not be healthy, in-situ, or death if $\tildeh$ is healthy. That is, $h = 1$, $i \in \bO$, $h' = 3$, $\tildeh=1$, $\ti \in \bO$, and $\tildeh' \in \{1,2,4,7\}$.
\item If $(h,i,h')$ equals (healthy, whatever emission, death), then the counterfactual state $\tildeh'$ can not be healthy, in-situ, or invasive if $\tildeh$ is healthy. That is, $h = 1$, $i \in \bO$, $h' = 7$, $\tildeh=1$, $\ti \in \bO$, and $\tildeh' \in \{1,2,3,4,5\}$.
\ei

\paragraph{State $h=2$ (undiagnosed in-situ).} We enforce PM for the following combinations:
\bi
\item If $(h,i,h')$ equals (in-situ, undetected,  in-situ), then the counterfactual state $\tildeh'$ can not be invasive or death if $\tildeh$ is healthy or in-situ. That is,  $h = 2$, $i \in \{1,2,3\}$, $h' = 2$, $\tildeh \in \{1,2,4\}$, $\ti \in \bO$, and $\tildeh' \in \{3,5,7\}$.
\item If $(h,i,h')$ equals (in-situ, detected,  in-situ), then the counterfactual state $\tildeh'$ can not be invasive or death if $\tildeh$ is in-situ and detected. That is,  $h = 2$, $i = 4$, $h' = 4$, $\tildeh \in \{2,4\}$, $\ti = 4$, and $\tildeh' \in \{3,5,7\}$.
\item If $(h,i,h')$ equals (in-situ,  undetected,  invasive), then the counterfactual state $\tildeh'$ can not be death if $\tildeh$ is healthy or in-situ. That is,  $h = 2$, $i \in \{1,2,3\}$, $h' = 3$, $\tildeh \in \{1,2,4\}$, $\ti \in \bO$, and $\tildeh' = 7$.
\item If $(h,i,h')$ equals (in-situ,  detected,  invasive), then the counterfactual state $\tildeh'$ can not be death if $\tildeh$ is in-situ and detected. That is,  $h = 2$, $i = 4$, $h' = 5$, $\tildeh \in \{2,4\}$, $\ti = 4$, and $\tildeh' = 7$.
\item If $(h,i,h')$ equals (in-situ,  detected,  recovered), then the counterfactual state $\tildeh'$ can not be in-situ, invasive, or death if $\tildeh$ is in-situ and detected. That is,  $h = 2$, $i = 4$, $h' = 6$, $\tildeh \in \{2,4\}$, $\ti = 4$, and $\tildeh' \in \{2,3,4,5,7\}$.
\item If $(h,i,h')$ equals (in-situ,  undetected,  death), then the counterfactual state $\tildeh'$ can not be in-situ, invasive, or recovered if $\tildeh$ is in-situ and undetected. That is,  $h = 2$, $i \in \{1,2,3\}$, $h' = 7$, $\tildeh = 2$, $\ti \in \{1,2,3\}$, and $\tildeh' \in \{2,3,4,5,6\}$.
\item If $(h,i,h')$ equals (in-situ,  detected,  death), then the counterfactual state $\tildeh'$ can not be in-situ, invasive, or recovered if $\tildeh$ is in-situ and detected. That is,  $h = 2$, $i = 4$, $h' = 7$, $\tildeh \in \{2,4\}$, $\ti = 4$, and $\tildeh' \in \{2,3,4,5,6\}$.
\ei

\paragraph{State $h=3$ (undiagnosed invasive).} We enforce PM for the following combinations:
\bi
\item If $(h,i,h')$ equals (invasive, undetected,  invasive), then the counterfactual state $\tildeh'$ can not be death if $\tildeh$ is healthy, in-situ, or invasive. That is,  $h = 3$, $i \in \{1,2,3\}$, $h' = 3$, $\tildeh \in \{1,2,3,4,5\}$, $\ti \in \bO$, and $\tildeh' = 7$.
\item If $(h,i,h')$ equals (invasive, detected,  invasive), then the counterfactual state $\tildeh'$ can not be death if $\tildeh$ is invasive and detected. That is,  $h = 3$, $i = 5$, $h' = 5$, $\tildeh \in \{3,5\}$, $\ti = 5$, and $\tildeh' = 7$.
\item If $(h,i,h')$ equals (invasive, detected,  recovered), then the counterfactual state $\tildeh'$ can not be invasive or death if $\tildeh$ is invasive and detected. That is,  $h = 3$, $i = 5$, $h' = 6$, $\tildeh \in \{3,5\}$, $\ti = 5$, and $\tildeh' \in \{3,5,7\}$.
\item If $(h,i,h')$ equals (invasive,  undetected,  death), then the counterfactual state $\tildeh'$ can not be invasive or recovered if $\tildeh$ is invasive and undetected. That is,  $h = 3$, $i \in \{1,2,3\}$, $h' = 7$, $\tildeh \in \{3,5\}$, $\ti \in \{1,2,3\}$, and $\tildeh' \in \{3,5,6\}$.
\item If $(h,i,h')$ equals (invasive,  detected,  death), then the counterfactual state $\tildeh'$ can not be invasive or recovered if $\tildeh$ is invasive and detected. That is,  $h = 3$, $i = 5$, $h' = 7$, $\tildeh \in \{3,5\}$, $\ti = 5$, and $\tildeh' \in \{3,5,6\}$.
\ei

\paragraph{State $h=4$ (diagnosed in-situ).} We enforce PM for the following combinations:
\bi
\item If $(h,i,h')$ equals (in-situ,  detected,  in-situ), then the counterfactual state $\tildeh'$ can not be invasive, recovered, or death if $\tildeh$ is in-situ and detected.  That is,  $h = 4$, $i = 4$, $h' = 4$, $\tildeh \in \{2,4\}$, $\ti = 4$, and $\tildeh' \in \{5,6,7\}$.
\item If $(h,i,h')$ equals (in-situ,  detected,  invasive), then the counterfactual state $\tildeh'$ can not be in-situ, recovered, or death if $\tildeh$ is in-situ and detected.  That is,  $h = 4$, $i = 4$, $h' = 5$, $\tildeh \in \{2,4\}$, $\ti = 4$, and $\tildeh' \in \{4,6,7\}$.
\item If $(h,i,h')$ equals (in-situ,  detected,  recovery), then the counterfactual state $\tildeh'$ can not be in-situ, invasive, or death if $\tildeh$ is in-situ and detected.  That is,  $h = 4$, $i = 4$, $h' = 6$, $\tildeh \in \{2,4\}$, $\ti = 4$, and $\tildeh' \in \{4,5,7\}$.
\item If $(h,i,h')$ equals (in-situ,  detected,  death), then the counterfactual state $\tildeh'$ can not be in-situ, invasive, or recovered if $\tildeh$ is in-situ and detected.  That is,  $h = 4$, $i = 4$, $h' = 7$, $\tildeh \in \{2,4\}$, $\ti = 4$, and $\tildeh' \in \{4,5,6\}$.
\ei

\paragraph{State $h=5$ (diagnosed invasive).} We enforce PM for the following combinations:
\bi
\item If $(h,i,h')$ equals (invasive,  detected,  invasive), then the counterfactual state $\tildeh'$ can not be recovered or death if $\tildeh$ is invasive and detected.  That is,  $h = 5$, $i = 5$, $h' = 5$, $\tildeh \in \{3,5\}$, $\ti = 5$, and $\tildeh' \in \{6,7\}$.
\item If $(h,i,h')$ equals (invasive,  detected,  recovery), then the counterfactual state $\tildeh'$ can not be invasive or death if $\tildeh$ is invasive and detected.  That is,  $h = 5$, $i = 5$, $h' = 6$, $\tildeh \in \{3,5\}$, $\ti = 5$, and $\tildeh' \in \{5,7\}$.
\item If $(h,i,h')$ equals (invasive,  detected,  death), then the counterfactual state $\tildeh'$ can not be invasive or recovery if $\tildeh$ is invasive and detected.  That is,  $h = 5$, $i = 5$, $h' = 7$, $\tildeh \in \{3,5\}$, $\ti = 5$, and $\tildeh' \in \{5,6\}$.
\ei

\paragraph{State $h=6$ (recovery).} No combination for which we enforce PM.

\paragraph{State $h=7$ (death).} No combination for which we enforce PM.

\subsection{Details on the Comonotonic Copula} \label{sec:comonocancer}

We discussed the counterfactual simulation under the comonotonic copula for a general dynamic latent-state model in \S\ref{sec:CopulaSimComon}. In this section, we connect that discussion to the breast cancer application.  To do so, it suffices to define the rank functions $r_{H}(\cdot)$ (for states) and $r_O(\cdot)$ (for emissions). For states, there are two possible orderings that seem ``natural'' (from ``best'' to ``worst''):
\bi
\item $(1, 6, 4, \textbf{2},  \textbf{5}, 3, 7)$
\item $(1, 6, 4, \textbf{5}, \textbf{2}, 3, 7)$.
\ei
Recalling the $\fQ(i)$ notation from \S\ref{sec:CancerHMMPrimitives},  observe that columns 2 and 5 are never ``active'' simultaneously in any row of $\fQ(i)$ (for any $i$).  Hence, the choice of ordering (between the two orderings above) will not matter and we can pick any one.  Suppose we pick the first one. Then, this ordering defines the rank function. For example, $r_H(6) = 2$, i.e., rank of state 6 equals 2.   For the inverse function,  $r_{H}^{-1}(2) = 6$.

It is unclear how to define $r_O(\cdot)$ for the breast cancer application but as it turns out,  it does not matter.  To see why, consider the generic path of interest (from \eqref{eq:path}):
\begin{align*}
\underbrace{o_1,  \ldots,  o_{\tau_s-1}}_{\in \{2,3\}},   \red{\underbrace{o_{\tau_s}, \ldots, o_{\tau_e}}_{=1}},  \underbrace{o_{\tau_e+1}, \ldots, o_{\tau_d-1}}_{\in \{4,5\}},  \underbrace{o_{\tau_d:T}}_{=7}.
\end{align*}
For the first $\tau_s-1$ periods, observe that the counterfactual emission $\tildeo_{1:\tau_s-1}(b)$ equals the observed emission $o_{1:\tau_s-1}$ for each $b$. This is because the intervention policy $\tx_{1:\tau_s-1}$ equals the observed policy $x_{1:\tau_s-1}$.  Now, consider periods $\tau_s$ to $\tau_e$, during which the screening was not done, i.e., $x_{\tau_s : \tau_e} = 1$.  Hence, the emissions $o_{\tau_s : \tau_e} = 1$ w.p.\ 1 (see the matrix $\fE(1)$ in \S\ref{sec:CancerPrimitiveE}). This means that the emissions does not contain any information regarding the underlying noise variables $V_{\tau_s : \tau_e}$ (see Figure \ref{fig:SCMComono}) and hence,  their posterior equals their prior, which is $\text{Unif}[0,1]$. As such,  for $t \in \{\tau_s, \ldots, \tau_e\}$, we can sample $\tildeo_t(b)$ using the categorical distribution over the probability vector $[e_{ \tildeh_t(b) \tx_t i}]_i$.  Note that we can use $\tildeo_{\tau_s-1}(b)$ to sample $\tildeh_{\tau_s}(b)$, which we can use to sample $\tildeo_{\tau_s}(b)$, and so on (until we have sampled $\tildeh_{\tau_e+1}(b)$).
Now, consider $t = \tau_e+1$.  We know $o_{t} \in \{4,5\}$:
\bi
\item If $o_t = 4$, then $h_t(b) = 2$ and $\tildeh_t(b) \in \{2,4,6\}$ (cf.\ pathwise monotonicity).
\bi
\item If $\tildeh_t(b) \in \{4,6\}$, then $\tildeo_{t}(b) = \tildeh_{t}(b)$ (since rows 4 and 6 of $\fE(0)$ have $1$ on the diagonal).
\item Else, if $\tildeh_t(b) = 2$ ($= h_{t}(b)$), then $\tildeo_{t}(b) = o_{t} =4$.
\ei
\item Else, if $o_{t} = 5$, then $h_{t}(b) = 3$ and $\tildeh_t(b) \in \{3,4,5,6\}$ (cf.\ pathwise monotonicity).
\bi
\item If $\tildeh_t(b) \in \{4,5,6\}$, then $\tildeo_{t}(b) = \tildeh_t(b)$ (since rows 4, 5,  6 of $\fE(0)$ have 1 on the diagonal).
\item If $\tildeh_t(b) = 3$ ($= h_{t}(b)$), then $\tildeo_{t}(b) = o_t = 5$.
\ei
\ei
Finally, for $t \ge \tau_e+2$,  we know $h_t(b) \ge 4$ and that the corresponding rows in $\fE(0)$ are 0-1. Hence, the posterior of $V_t$ equals the prior and we can sample $\tildeo_t(b)$ using the categorical distribution over the probability vector $[e_{ \tildeh_t(b) \tx_t i}]_i$.
By construction,  the comonotonic copula will obey pathwise monotonicity and hence, will ensure that in the counterfactual world,  patient does not die before period $T$, i.e., $\tH_{T-1} \neq 7$ w.p.\ 1.

\subsection{Results for Path 2} \label{sec:pathtworesults}

\begin{figure}[H]
  \begin{center}
    \subfigure[UB / LB]{\label{fig:PN4a}\includegraphics[width=.3\linewidth]{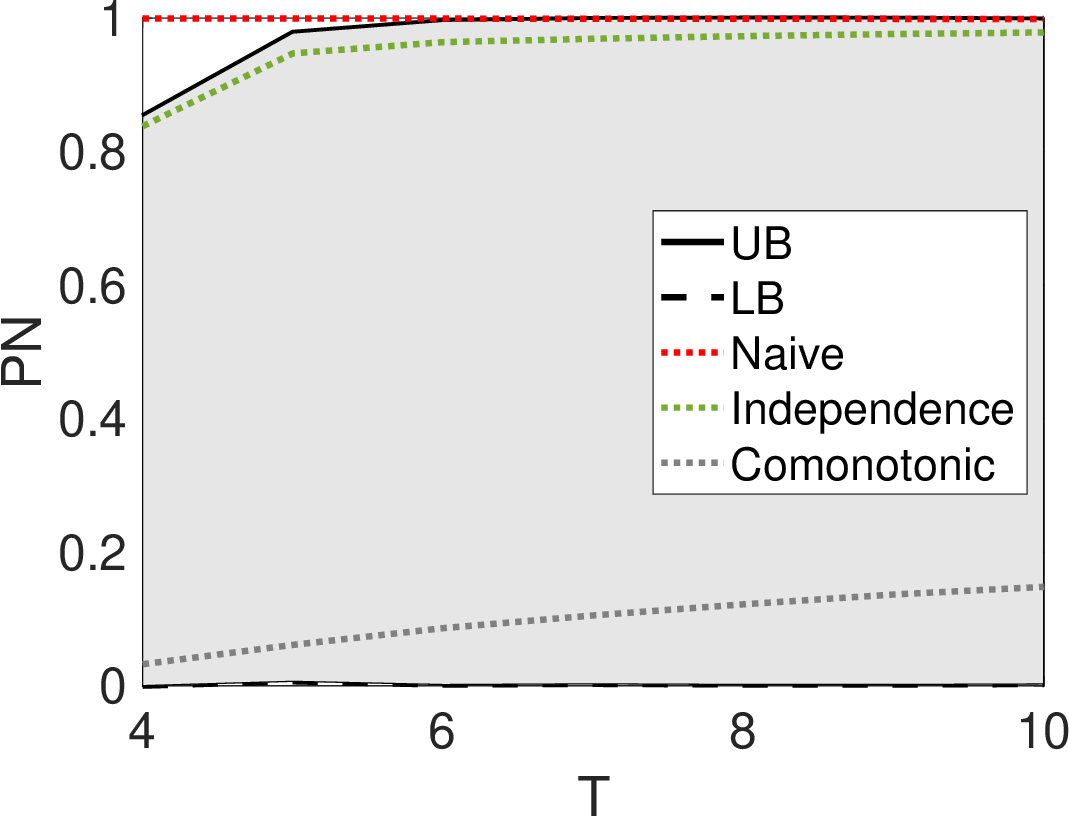}}
    \quad
    \subfigure[UB / LB with CS]{\label{fig:PN4b}\includegraphics[width=.3\linewidth]{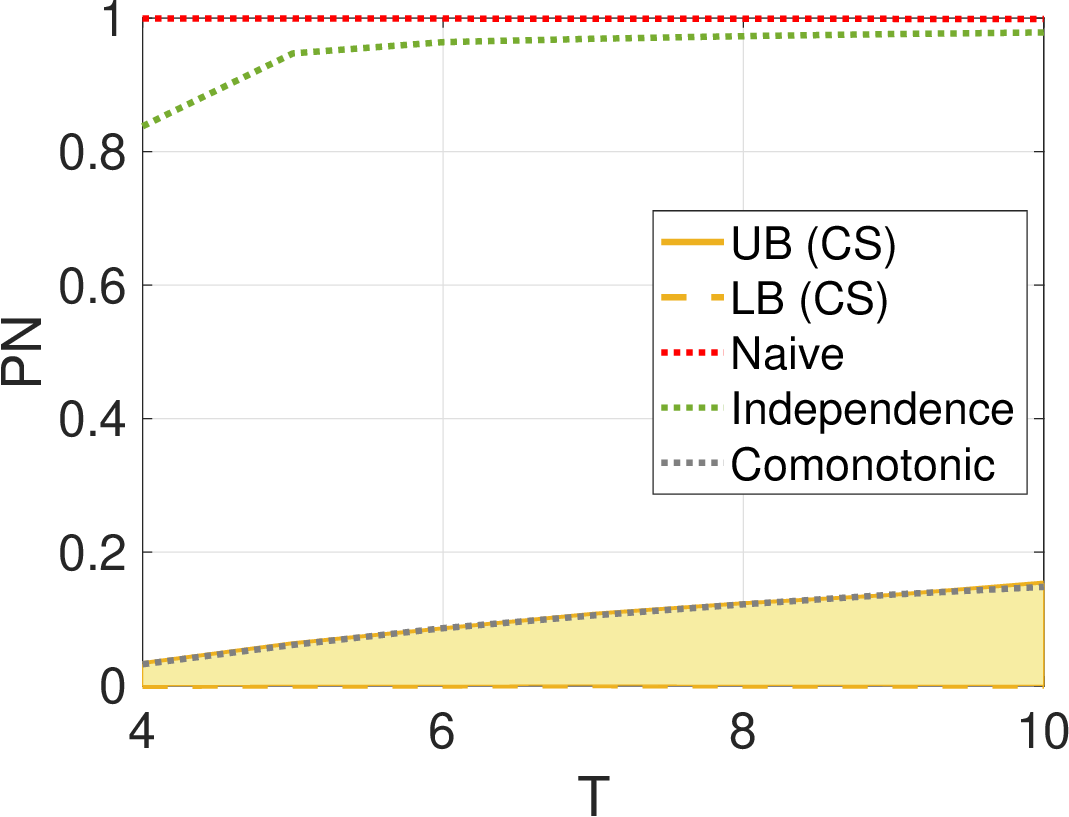}}
    \quad
    \subfigure[UB / LB with PM]{\label{fig:PN4c}\includegraphics[width=.3\linewidth]{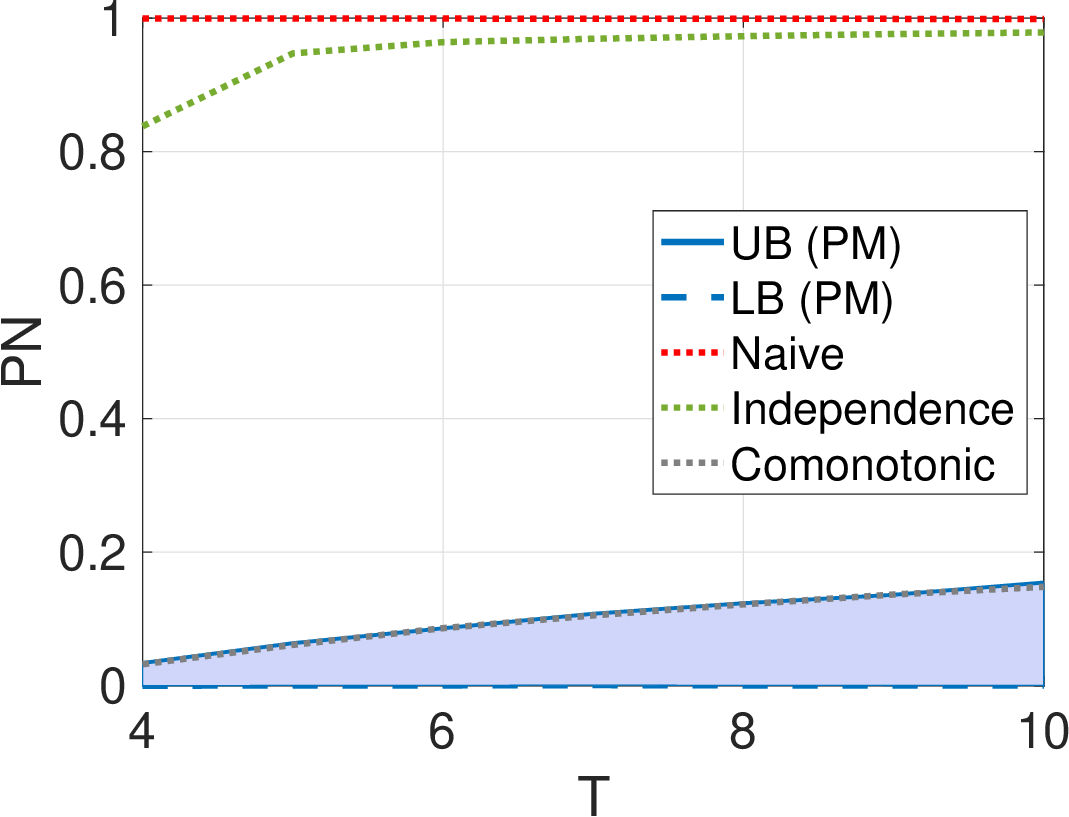}}
  \end{center}
\caption{PN results for path 2 as we vary $T \in \{4,\ldots,10\}$ (analogous to Figure \ref{fig:PN3} for path 1).
As in Figure \ref{fig:PN3}, observe that the \texttt{LB}, \texttt{LB(CS)}, and \texttt{LB(PM)} curves coincide (the lowest curve in each figure).
Further, the \texttt{UB(CS)} and \texttt{UB(PM)} curves coincide for path 2. }
\label{fig:PN4}
\end{figure}

\end{document}

%% file: definitions.tex
\def\red#1{{\color{red}#1}}

\newcommand\leaveline{\vspace{0.3cm}}

\def\bi{\begin{itemize}}
\def\ei{\end{itemize}}

\newcommand{\bI}{\mathbb{I}}

\newcommand{\bO}{\mathbb{O}}
\newcommand{\bH}{\mathbb{H}}
\newcommand{\bX}{\mathbb{X}}
\newcommand{\bP}{\mathbb{P}}
\newcommand{\bE}{\mathbb{E}}

\newcommand{\cF}{\mathcal{F}}

\newcommand{\CP}{C^{\mbox{\scriptsize P}}}
\newcommand{\CN}{C^{\mbox{\scriptsize N}}}
\newcommand{\CI}{C^{\mbox{\scriptsize I}}}

\newcommand{\PNUB}{\text{PN}^{\mbox{\scriptsize ub}}}
\newcommand{\PNLB}{\text{PN}^{\mbox{\scriptsize lb}}}
\newcommand{\PNUBCS}{\text{PN}^{\mbox{\scriptsize ub}}_{\mbox{\scriptsize cs}}}
\newcommand{\PNLBCS}{\text{PN}^{\mbox{\scriptsize lb}}_{\mbox{\scriptsize cs}}}
\newcommand{\PNUBPM}{\text{PN}^{\mbox{\scriptsize ub}}_{\mbox{\scriptsize pm}}}
\newcommand{\PNLBPM}{\text{PN}^{\mbox{\scriptsize lb}}_{\mbox{\scriptsize pm}}}

\newcommand{\Pb}{\mathbb{P}}
\newcommand{\fQ}{\mathbf{Q}}
\newcommand{\fE}{\mathbf{E}}
\newcommand{\tfQ}{\mathbf{\widetilde{Q}}}
\newcommand{\tfE}{\mathbf{\widetilde{E}}}

\newcommand{\fU}{\mathbf{U}}
\newcommand{\fM}{\mathbf{M}}
\newcommand{\fV}{\mathbf{V}}

\newcommand{\fp}{\mathbf{p}}
\newcommand{\tfp}{\mathbf{\widetilde{p}}}

\newcommand{\ftheta}{\pmb{\theta}}
\newcommand{\fpi}{\bm{\pi}}

\newcommand{\tH}{\widetilde{H}}
\newcommand{\tildeh}{\widetilde{h}}
\newcommand{\tildeo}{\widetilde{o}}

\newcommand{\tx}{\widetilde{x}}

\newcommand{\ti}{\widetilde{i}}
\newcommand{\te}{\widetilde{e}}
\newcommand{\tq}{\widetilde{q}}
\newcommand{\tM}{\widetilde{\fM}}

\newcommand{\tY}{\widetilde{Y}}

\newcommand{\bPi}{\bm{\Pi}}

\newcommand{\q}{\mathsf{q}}
\newcommand{\e}{\mathsf{e}}

\newtheorem{defi}{Definition}
\newtheorem{thm}{Theorem}
\newtheorem{prop}{Proposition}

\newtheorem{ex}{Example}
\newtheorem{rem}{Remark}